\documentclass[10pt,twocolumn,letterpaper]{article}

\usepackage[pagenumbers]{cvpr} % To force page numbers, e.g. for an arXiv version

\usepackage{amsmath}
\usepackage{multirow}
\usepackage{pifont}
\usepackage{makecell}
\usepackage{subcaption}
\usepackage{colortbl} % For command \arrayrulecolor

\newcommand{\methodname}[1]{\texttt{UNITE}}
\definecolor{cvprblue}{rgb}{0.21,0.49,0.74}
\usepackage[pagebackref,breaklinks,colorlinks,allcolors=cvprblue]{hyperref}

\def\paperID{****} % *** Enter the Paper ID here
\def\confName{CVPR}
\def\confYear{2025}

\title{Towards a Universal Synthetic Video Detector: From Face or Background Manipulations to Fully AI-Generated Content\thanks{This paper has been published at The IEEE/CVF Conference on Computer Vision and Pattern Recognition 2025}}
\author{Rohit Kundu$^{1, 2}$, Hao Xiong$^{1}$, Vishal Mohanty$^{1}$, Athula Balachandran$^{1}$, Amit K. Roy-Chowdhury$^{2}$\\
$^{1}$Google, Mountain View, USA ~$^{2}$University of California, Riverside\\
{\tt \small \{rohitkun, haoxg, vishalmohanty, athula\}@google.com, amitrc@ece.ucr.edu}}
\begin{document}
\maketitle
\begin{abstract}
    Existing DeepFake detection techniques primarily focus on facial manipulations, such as face-swapping or lip-syncing. However, advancements in text-to-video (T2V) and image-to-video (I2V) generative models now allow fully AI-generated synthetic content and seamless background alterations, challenging face-centric detection methods and demanding more versatile approaches.

    To address this, we introduce the \underline{U}niversal \underline{N}etwork for \underline{I}dentifying \underline{T}ampered and synth\underline{E}tic videos (\methodname{}) model, which, unlike traditional detectors, captures full-frame manipulations. \methodname{} extends detection capabilities to scenarios without faces, non-human subjects, and complex background modifications. It leverages a transformer-based architecture that processes domain-agnostic features extracted from videos via the SigLIP-So400M foundation model. Given limited datasets encompassing both facial/background alterations and T2V/I2V content, we integrate task-irrelevant data alongside standard DeepFake datasets in training. We further mitigate the model’s tendency to over-focus on faces by incorporating an attention-diversity (AD) loss, which promotes diverse spatial attention across video frames. Combining AD loss with cross-entropy improves detection performance across varied contexts. Comparative evaluations demonstrate that \methodname{} outperforms state-of-the-art detectors on datasets (in cross-data settings) featuring face/background manipulations and fully synthetic T2V/I2V videos, showcasing its adaptability and generalizable detection capabilities. 
\end{abstract}    
\section{Introduction} \label{sec:intro}

\begin{figure}
    \centering
    \includegraphics[width=0.9\columnwidth]{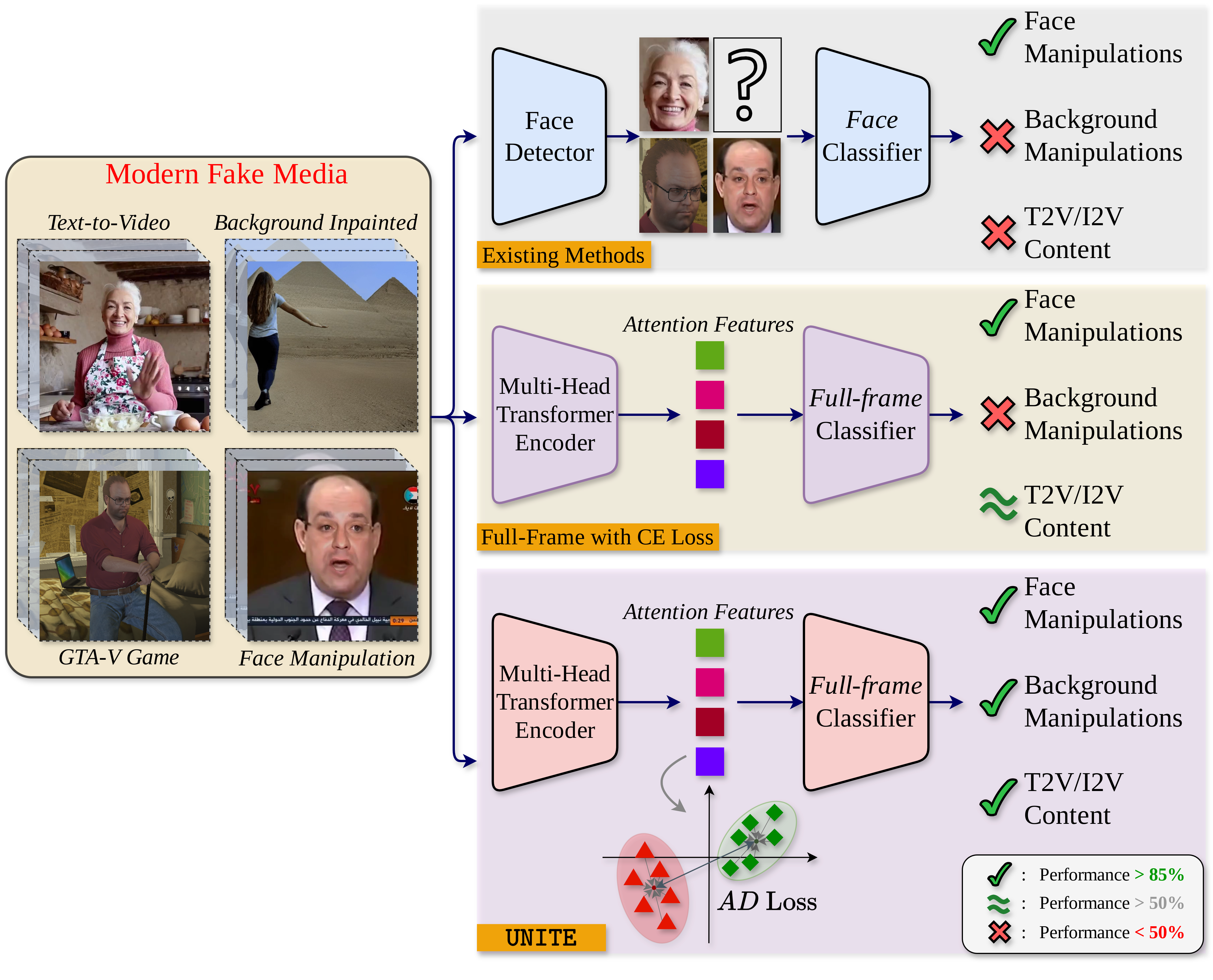}\vspace{-1em}
    \caption{\textbf{Problem Overview}: Existing DeepFake detection methods primarily focus on identifying face-manipulated videos, most of which cannot perform inference unless there is a face detected in the video. However, with advancements like seamless background modifications (e.g., AVID \cite{zhang2024avid}) and hyper-realistic content from games like GTA-V \cite{hu2021sail} and T2V/I2V models \cite{chen2024demamba}, a more comprehensive approach is needed. A model trained with only cross-entropy (CE) loss, using full frames, automatically focuses on the face, capturing temporal discontinuities through its transformer architecture, performing better than random ($\approx$) on T2V/I2V content but struggling with background manipulations. \methodname{}, with its attention-diversity (AD) loss, effectively detects both face/background manipulations and fully synthetic content.}
    \vspace{-1.5em}
    \label{fig:problems_statement}
\end{figure}

The rise of synthetic media, especially DeepFakes, has transformed content perception and interaction online. Hyper-realistic images produced by technologies like FLUX$_{1.1}$ \cite{fluxgenerator} are challenging for even humans to identify as fake, and with the development of their video model underway, this underscores the critical need for effective detection methods for media generated by deep neural networks. Traditional DeepFake generators \cite{li2019faceshifter, thies2019deferred} focus on manipulating human faces through face-swapping and lip-syncing. However, powerful text-to-video (T2V) and image-to-video (I2V) models \cite{wang2023modelscope, runwayresearch, opensora} have expanded manipulation possibilities beyond faces.

While conventional detectors \cite{cheng2024can, sun2024diffusionfake, xu2023tall} perform well on older, face-centric DeepFake datasets \cite{dolhansky2020deepfake, li2020celeb, rossler2019faceforensics++}, they often struggle with newer manipulations involving full scenes or backgrounds. DeepFake-O-Meter \cite{ju2024deepfake} is an open-source tool (consisting of several state-of-the-art models) for DeepFake detection, but it cannot run inference unless a human face is visible in the image/video. The rapid spread of misinformation, particularly during critical periods such as elections, highlights the need for generalizable detection models capable of identifying diverse manipulations, including face, background, and fully AI-generated T2V/I2V content with/without human subjects (overview in Fig. \ref{fig:problems_statement}).

To address these challenges, we present a \underline{U}niversal \underline{N}etwork for \underline{I}dentifying \underline{T}ampered and synth\underline{E}tic videos or \methodname{} model, which detects both partially manipulated (foreground/background) and fully synthetic videos. Unlike detectors focused solely on face detection, \methodname{} analyzes entire frames, regardless of whether a human subject is present in the videos. 

Given the inherent domain gaps in DeepFake datasets \cite{seraj2024semi, zhou2024fine}, even when generated by similar techniques, we leverage the SigLIP-So400m \cite{alabdulmohsin2024getting} foundation model to extract domain-agnostic features. This serves as inputs to a learnable transformer with multi-head attention, enabling effective detection by capturing temporal inconsistencies in synthetic content. However, preliminary experiments show that training a transformer architecture solely with cross-entropy (CE) loss often leads to the attention heads converging on the face region (Fig. \ref{fig:heatmaps}). As a result, the model struggles during inference when handling videos where a real human subject is placed in a manipulated background or content generated by T2V \cite{wang2023modelscope, chen2023videocrafter1} and I2V \cite{ma2024latte, opensora} models. To address this, we introduce an ``attention-diversity" (AD) loss that encourages attention heads to focus on different spatial regions, enhancing the model's ability to capture critical cues from both foreground and background.

Due to the limited availability of open-source datasets covering face or background manipulations and fully AI-generated content, we employ innovative training strategies for \methodname{}. This includes integrating task-irrelevant data with standard DeepFake datasets to simulate AI-generated synthetic media. Beyond the popularly used FaceForensics++ \cite{rossler2019faceforensics++} dataset, we utilize the SAIL-VOS-3D \cite{hu2021sail} dataset, originally designed for 3D video object segmentation in the game GTA-V. As this dataset is fully synthetic, it helps simulate AI-generated media, enhancing our model's ability to detect diverse forms of synthetic manipulation.

Our contributions can be summarized as follows:

\begin{itemize}
    \item We propose \methodname{}, a model for detecting partially manipulated (foreground/background) and fully synthetic videos, moving beyond face-centric DeepFake detection.
    % \amit{Need to add that it also detects facial deepfakes without loss of performance.}\rohit{We mentioned that we can detect partially manipulated content too.}
    
    \item Unlike detectors relying on face detection and cropping, our model can process full video frames and can detect fakes even if there are no human subjects. 
    % \amit{Is this really a contribution?} \rohit{Yes, since existing detectors cannot work unless there are faces to detect, e.g., (\href{https://zinc.cse.buffalo.edu/ubmdfl/deep-o-meter/}{DeepFake-o-meter} runs into error if there are no human faces in the video). We can do fake media detection even if humans are not present}
    
    \item Using the SigLIP-So400m foundation model \cite{alabdulmohsin2024getting}, we extract domain-agnostic features, enabling generalization to in-the-wild DeepFakes.
    
    \item We introduce an attention-diversity loss, encouraging attention heads to focus on diverse spatial regions, enhancing detection beyond face manipulations.
    
    % \item To address the lack of open-source datasets containing both partially manipulated and fully AI-generated video, we integrate task-irrelevant data (e.g., SAIL-VOS-3D \cite{hu2021sail}) improving detection of varied synthetic content.
    % \amit{why did you comment out the previous bullet about training strategies?} \rohit{Since it is more of a strategy than a contribution (as we did not create the data ourselves, we just used an existing dataset), I removed it to save space.}
    
    \item Unlike existing methods that evaluate on a few datasets, we comprehensively assess \methodname{} on a broad range of face and synthetic datasets with various T2V/I2V generators, outperforming detection in foreground, background, and T2V/I2V manipulations.

\end{itemize}
  
\section{Related Work} \label{sec:rel_work}
% The existing state-of-the-art DeepFake detection methods can be broadly categorized into methods that model visual artifacts within video frames and methods that model the temporal inconsistencies across frames.

\noindent
\textbf{Face-centric DeepFake Detection:} Methods such as \cite{durall2019unmasking, concas2024quality, pellicer2024pudd} focus on modeling spatial inconsistencies for DeepFake detection. Concas et al. \cite{concas2024quality} track specific facial features (eyes, nose, mouth) in a high-frequency domain, as fake videos exhibit distinct behaviors in this space compared to real ones. However, due to their dataset-centric architectures, it is not a generalizable approach. PUDD \cite{pellicer2024pudd} learns person-specific prototypes by modeling patterns from real videos and assessing how closely inference videos match these prototypes, making it less effective for detecting in-the-wild DeepFakes. LAA-Net \cite{nguyen2024laa} attempts to improve generalizability through a multi-task attention module that detects artifacts via heatmap and self-consistency regression, but it works by analyzing every still frame of a video, thus limiting its scalability. Mazaheri et al. \cite{mazaheri2022detection} tackled the challenging problem of detecting and localizing expression swaps, where the number of manipulated pixels is even fewer compared to identity swaps. Although the authors obtained good localization performance, their reliance on a CNN-based architecture limits their ability to capture temporal inconsistencies effectively. 

DPNet \cite{trinh2021interpretable} and ID-Reveal \cite{cozzolino2021id} are identity-aware temporal artifact modeling methods: DPNet \cite{trinh2021interpretable} focuses on interpretable, prototype-based detection of unnatural movements through dynamic feature representations, while ID-Reveal \cite{cozzolino2021id} leverages metric learning on real data to detect biometric inconsistencies. However these methods assume access to authentic reference videos, making them unfit for in-the-wild DeepFake detection. Shifting from identity-specific methods, TALL \cite{xu2023tall} constructs composite ``thumbnail" images from four consecutive frames taken at random timestamps for temporal analysis. TI2Net \cite{liu2023ti2net} captures temporal anomalies by subtracting consecutive frame features but relies on the assumption that FaceSwap manipulations show pronounced inter-frame discrepancies.

Choi et al. \cite{choi2024exploiting} approach DeepFake detection by targeting suppressed variance in style-based latent temporal features via a StyleGRU module. However, this method is limited to cropped face regions, making it ineffective for T2V/I2V or background-manipulated videos.

\begin{figure}
    \centering
    \includegraphics[width=0.8\columnwidth]{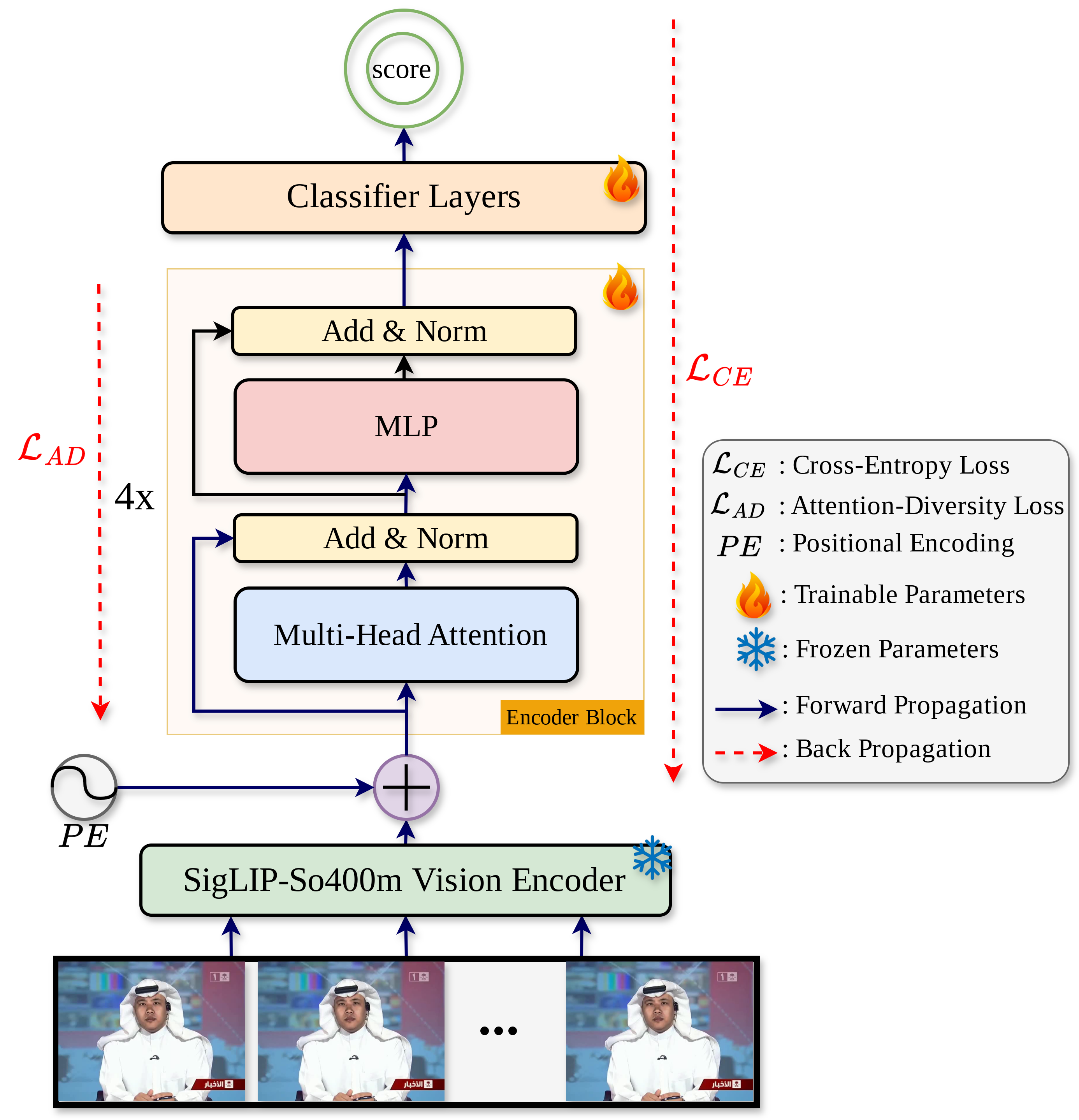} \vspace{-1em}
    \caption{\textbf{\methodname{} architecture overview:} We extract domain-agnostic features ($\xi$) using the SigLIP-So400m foundation model \cite{alabdulmohsin2024getting} to mitigate domain gaps between DeepFake datasets (Sec. \ref{subsec:siglip}). These embeddings, combined with positional encodings, are input to a transformer with multi-head attention and MLP layers (Sec. \ref{subsec:transformer}), culminating in a classifier for final predictions. AD-loss (Sec. \ref{subsec:adloss}) encourages the model attention to span diverse spatial regions.}
    \label{fig:overall}
    \vspace{-1em}
\end{figure}

\noindent
\textbf{Synthetic Video Detection:} While there has been considerable effort in synthetic image detection \cite{corvi2023detection, corvi2023intriguing, wang2023dire}, synthetic video detection is a scarcely explored area. DeMamba \cite{chen2024demamba}, a recently proposed synthetic video detector, analyzes small zones of video frames to capture how pixels change spatiotemporally. Its continuous scanning approach better tracks subtle changes and patterns, making it easier to spot manipulated or fake content. The authors also propose a million scale T2V/I2V dataset (which we call the DeMamba dataset hereforth) to evaluate their performance. However, the method was not evaluated on typical DeepFake datasets \cite{rossler2019faceforensics++, li2020celeb, wang2021hififace} and thus is not fit for human-centric DeepFake detection. In contrast our \methodname{} model is built to detect partially manipulated as well as fully AI-generated videos. A key reason for the scarcity of synthetic video detection methods in the literature is the lack of standardized datasets for training and evaluation.
\section{Proposed Method} \label{sec:method}
In this section we first describe our problem setup (Sec. \ref{subsec:data}), then we describe the foundation model-based video encoding (Sec. \ref{subsec:siglip}), our trainable transformer architecture (Sec. \ref{subsec:transformer}) and finally the novel loss function for training the \methodname{} model (Sec. \ref{subsec:adloss}). The overview of the \methodname{} architecture is shown in Fig. \ref{fig:overall}.

%-----------------------------------------------------------
\subsection{Problem Setup}\label{subsec:data}
For each video in the dataset, we implement a frame sampling strategy where every other frame is extracted. From the resulting collection of frames, we construct video segment samples consisting of $n_f = 64$ consecutive frames, which are defined as a single data sample, denoted as ``$v$". This selection aligns with the context window established for our video transformer model, ensuring that each sample retains the temporal coherence necessary for effective training. In instances where the video duration results in fewer than $n_f$ frames, we apply padding (ablation in supplementary) to ensure that each sample remains uniform in size.

% Given that the original video-level labels (``real" or ``fake") are known and fake videos contain manipulations throughout their entirety (i.e., there are no videos with manipulations limited to specific frames), each $n_f$-frame segment generated from a given video is assigned the same label (denoted by $l \in \{0,\hdots, n_c\}$) as the source video with $n_c$ being the number of possible classes. The aggregation of these samples across all the video files forms our final video dataset, $\{v_i, l_i\}_{i=1}^{N} \in \mathcal{V}$, where $N$ is the total number of samples obtained after this procedure. This approach effectively acts as a data augmentation technique, as it increases the number of training samples without additional data collection.

Given that the original video-level classification labels are known and fake videos contain manipulations throughout their entirety (i.e., no videos have manipulations limited to specific frames), each $n_f$-frame segment generated from a given video is assigned the same label (denoted by $l \in {0,\hdots, n_c}$) as the source video, with number of possible classes as $n_c$. The aggregation of these samples across all videos forms our final dataset, ${v_i, l_i}_{i=1}^{N} \in \mathcal{V}$, where $N$ is the total number of samples obtained. This approach serves as a data augmentation technique, increasing the number of training samples without additional data collection.

%-----------------------------------------------------------
\subsection{Domain-Agnostic Feature Extraction}\label{subsec:siglip}
Our objective is to leverage the trained \methodname{} model for detecting in-the-wild DeepFakes, necessitating careful consideration of the domain gap \cite{chen2021featuretransfer, lv2024domainforensics} between the train and test datasets. To effectively mitigate the impact of domain discrepancies, we utilize the SigLIP foundation model \cite{zhai2023sigmoid} to extract domain-agnostic features. Specifically, we employ the "shape-optimized" ViT \cite{alabdulmohsin2024getting} image encoder, whose architecture is rooted in \cite{dosovitskiy2020image}. The SigLIP-So400m \cite{alabdulmohsin2024getting} model was contrastively pretrained on 3B diverse examples using a sigmoid loss, enabling it to capture robust, generalizable features while maintaining a compact size (400M parameters). This extensive pretraining makes SigLIP particularly adept at extracting domain-invariant representations, crucial for handling the significant variability encountered in DeepFake datasets and real-world scenarios.

For every frame $f_j$ (resized to $\mathbb{R}^{384\times 384\times 3}$), of a video sample $v_i \in \mathcal{V}$, we obtain an image encoding from the frozen SigLIP-So400m \cite{alabdulmohsin2024getting} model as $e_j = SigLIP(f_j) \in \mathbb{R}^{t_s\times d_s}$, where $j \in \{1,2,\hdots,n_f\}$, since there are $n_f$ frames per video sample (as explained in Sec. \ref{subsec:data}), $t_s=729$ represents the token length and $d_s=1152$ represents the feature dimension. We concatenate these frame-level features for the video sample $v_i$, while maintaining the order of the frames, to obtain a video segment encoding denoted by $\xi_i \in \mathbb{R}^{n_f\times t_s \times d_s}$, which are used as input to our trainable transformer architecture. This encoded video dataset can be denoted as $\{\xi_i, l_i\}_{i=1}^N \in \mathcal{V}_{\xi}$.

%-----------------------------------------------------------
\subsection{Learnable Transformer Architecture}\label{subsec:transformer}
The \methodname{} architecture is a multi-head self-attention (MHSA) based transformer model designed specifically for video classification tasks. It leverages the strengths of Transformer networks to process sequences of video frame embeddings effectively, allowing for robust classification performance. 
The \methodname{} architecture allows for adjustable depth, signifying the number of encoder blocks stacked within the model. A deeper architecture can capture more intricate features and dependencies at the cost of increased computational complexity. For our architecture the depth is set to $4$. Ablation experiments with different transformer depths are provided in Sec. \ref{subsec:ablation}.

\subsubsection{Encoder Block}\label{subsubsec:encoder}
Each encoder block consists of the following components:

\noindent
\textbf{Multi-Head Self-Attention Layer:} This layer allows the model to focus on different parts of the input sequence simultaneously. With 12 attention heads in our architecture, it captures diverse interactions and dependencies among frames. Each head computes attention scores independently, learning multiple representations of the input data. This enhances the model's ability to detect subtle temporal variations, which is crucial for video classification. The use of scaled dot-product attention further helps mitigate large gradients and stabilizes training.

\noindent
\textbf{Layer Normalization and Residual Connections:} To aid gradient flow during training, each sub-layer incorporates residual connections that add the input back to the output, mitigating vanishing gradient issues in deep networks. Layer normalization follows to stabilize activations and enhance convergence speed. Additionally, dropout is applied post-attention to reduce overfitting by randomly deactivating a portion of neurons during training.

\noindent
\textbf{Feed-Forward Network (MLP):} The second sub-layer is a point-wise feed-forward network with two dense layers separated by a GELU activation \cite{hendrycks2016gaussian}. This MLP enables non-linear feature transformations, enhancing the model's expressiveness. By projecting attention outputs into a higher-dimensional space before reducing them, it captures complex interactions beyond linear transformations.

% The second sub-layer consists of a point-wise feed-forward network, which consists of two dense layers separated by a GELU activation function \cite{hendrycks2016gaussian}. This MLP allows for non-linear transformations of the features, enhancing the model's expressive power. By projecting the attention outputs into a higher-dimensional space before reducing them back, the MLP captures complex interactions that might not be evident through linear transformations alone.\\

% \noindent
Attention maps are derived from the outputs of the encoder blocks to be used for computing the Attention-Diversity loss (Sec. \ref{subsec:adloss}) and to illustrate how the model prioritizes different spatial regions in the frames during the classification process (Fig. \ref{fig:heatmaps}). This interpretability is valuable for understanding the model's decision-making, particularly in complex video classification scenarios.

\subsubsection{Positional Encoding}\label{subsubsec:positional_encoding}
We utilize a sine-cosine positional encoding scheme following the original transformer \cite{vaswani2017attention} to provide a unique position identifier for each input token. For every frame $f_j$ in the video sample with encoded feature dimension $d_s$ (as per Sec. \ref{subsec:siglip}), we compute the positional encoding for odd and even indexed feature dimensions as,

\begin{equation}\small
PE_{(j, 2i+1)} = \cos \left( \frac{j}{10000^{\frac{2i}{d_s}}} \right), \quad PE_{(j, 2i)} = \sin \left( \frac{j}{10000^{\frac{2i}{d_s}}} \right),
\label{eqn:PE}
\end{equation}
%
% \begin{equation}\label{eqn:odd_PE}
% PE_{(j, 2i+1)} = \cos \left( \frac{j}{10000^{\frac{2i}{d_s}}} \right) \text{, and}
% \end{equation}
%
% \begin{equation}\label{eqn:even_PE}
% PE_{(j, 2i)} = \sin \left( \frac{j}{10000^{\frac{2i}{d_s}}} \right),
% \end{equation}
where, $i$ denotes the feature index in the encoded feature dimension $d_s$. We embed these encodings directly into the input tokens of our video transformer to provide positional context, for better temporal modeling during the attention computation with minimal computational overhead.

%-----------------------------------------------------------
\subsection{Attention-Diversity Loss}\label{subsec:adloss}
Training a multi-head attention transformer with only cross-entropy loss makes the attention maps focus only on the face regions of the frame (as evident in Fig. \ref{fig:heatmaps}). However, we aim to detect fake videos where the manipulation might not be in the face at all, and instead be in the background (like video inpainting with the AVID model \cite{zhang2024avid}). To ensure that the attention heads focus on diverse spatial regions of the video frames, we devise the ``Attention-Diversity" (AD) loss.

AD-loss is designed to minimize overlap among attention maps (obtained from the first encoder block of the trainable transformer architecture in \ref{subsubsec:encoder}) while maintaining consistency across different input video samples. The attention outputs from the video transformer model $\mathcal{A}\in \mathbb{R}^{n_h\times t_s \times d_s}$ are used to pool the input SigLIP-So400m features $\xi \in \mathbb{R}^{n_f\times t_s \times d_s}$, resulting in a pooled feature tensor $\mathcal{P} \in \mathbb{R}^{n_h\times n_f}$ according to,

\begin{equation}
    \mathcal{P} = \sum_{j=1}^{t_s} \sum_{k=1}^{d_s} \mathcal{A}_{h, j, k} \cdot \xi_{f, j, k},
\end{equation}
where $f$ indexes the frames ($n_f$), $h$ indexes the number of attention heads ($n_h$), and $j, k$ are spatial positions.

We compute ``feature centers" which are points in the feature space that serve as anchors for the learned representations of different classes. These centers represent the average features of samples from a particular class, allowing the model to capture essential characteristics of that class. In the context of the AD-loss, feature centers $\mathcal{C} \in \mathbb{R}^{n_h \times n_f}$ are dynamically updated in each training iteration $\tau$ based on the pooled feature vectors $\mathcal{P}$, following,

\begin{equation}
    \mathcal{C}^{\tau} = \mathcal{C}^{\tau-1} - \eta \left( \mathcal{C}^{\tau-1} - \frac{1}{B} \sum_{b=1}^{B} \mathcal{P}_{b} \right),
\end{equation}
where $\eta$ represents the learning rate (set to $0.05$ in our experiments) for feature center updates and $B$ is the batch dimension. The feature centers are initialized with zeros ($\mathcal{C}^0 = [0]_{n_h\times n_f}$) at the beginning of the model training.

AD-loss consists of two distinct components: within-class loss and between-class loss inspired by the Fisher Discriminant Analysis for deep networks \cite{hanselmann2017deep}. The within-class term calculates the distance between the pooled feature vectors $\mathcal{P}$ and their corresponding feature centers $\mathcal{C}$, promoting closeness among similar samples. The between-class term measures the distance between feature centers of different classes, encouraging them to be spaced apart in the feature space, encouraging separability.

The within-class loss term is calculated as

\begin{equation}
    \mathcal{L}_{\text{within}} = \max\left( \left\| \mathcal{P} - \mathcal{C} \right\|_2 - \delta_{\text{within}}, 0 \right),
\end{equation}
where, $\delta_{\text{within}} \in \mathbb{R}^{n_c}$ is a hyperparameter controlling the allowable distance between feature vectors and their respective feature centers for inputs of the same class, encouraging tighter clustering and $\left\|\cdot \right\|_2$ represents $L_2$ normalization.

The between-class loss is calculated as

\begin{equation}
    \mathcal{L}_{\text{between}} = \sum_{\substack{k \neq l \\ (k,l) \in (n_h, n_h)}} \max\left( \delta_{\text{between}} - \left\| \mathcal{C}_{k} - \mathcal{C}_{l} \right\|_2, 0 \right),
\end{equation}
where $\delta_{\text{between}}$ is a predefined hyperparameter that ensures a minimum distance between feature centers of different classes. Thus the AD-loss is a simple addition of these two components $\mathcal{L}_{AD} = \mathcal{L}_{\text{within}} + \mathcal{L}_{\text{between}}$, making the final objective function for training the \methodname{} model,

\begin{equation}
    \mathcal{L}_{\methodname{}} = \lambda_1 \cdot \mathcal{L}_{CE} + \lambda_2 \cdot \mathcal{L}_{AD},
\end{equation}
where $\mathcal{L}_{CE}$ is the traditional cross-entropy loss, and $\lambda_1$ and $\lambda_2$ are loss-weighting hyperparameters.

%-----------------------------------------------------------  
\section{Experiments} \label{sec:experiments}

%-----------------------------------------------------------
\subsection{Datasets} \label{subsec:datasets}
% Most existing DeepFake datasets focus on face manipulations, such as face-swaps or lip-syncing, and lack fully AI-generated video content. While the DF40 dataset \cite{yan2024df40} has been introduced, its video data remains unavailable. The DeMamba dataset \cite{chen2024demamba} includes videos generated by text-to-video (T2V) and image-to-video (I2V) models, including diffusion models like SORA by OpenAI \cite{videoworldsimulators2024} and transformer models like Latte \cite{ma2024latte}. However, it primarily features natural scenes, limiting its relevance for training detectors aimed at identifying misinformation about individuals. We do use the validation split of DeMamba \cite{chen2024demamba} to evaluate \methodname{}.

% To train \methodname{}, we utilize the FaceForensics++ (FF++) \cite{rossler2019faceforensics++} dataset and the SAIL-VOS-3D dataset \cite{hu2021sail}, which comprises fully synthetic GTA-V game videos, mostly featuring human subjects suitable for DeepFake detection tasks. Additionally, results from training \methodname{} using FF++ and DeMamba are presented in the supplementary document.

Most DeepFake datasets primarily focus on face manipulations, such as face-swaps or lip-syncing, and lack fully AI-generated video content. The DeMamba dataset \cite{chen2024demamba} features videos from T2V/I2V models, including powerful diffusion models like SORA \cite{videoworldsimulators2024}. It is not a human-centric dataset and primarily contains videos from natural scenes. We use its validation split to evaluate \methodname{}.

To train \methodname{}, we employ the FaceForensics++ (FF++) \cite{rossler2019faceforensics++} dataset (c23 i.e., medium-quality) and the SAIL-VOS-3D dataset \cite{hu2021sail}, which includes synthetic (although not AI-generated) GTA-V game videos featuring human subjects suitable for DeepFake detection. To evaluate the \methodname{} model, we use:

\begin{itemize}
    \item \textbf{Face manipulated data:} FF++ \cite{rossler2019faceforensics++} (in-domain evaluation), CelebDF \cite{li2020celeb}, DeeperForensics \cite{jiang2020deeperforensics}, DeepfakeTIMIT \cite{korshunov2018deepfakes}, HifiFace \cite{wang2021hififace}, UADFV \cite{yang2019exposing}.
    \item \textbf{Background manipulated data:} Sample videos from the AVID \cite{zhang2024avid} model provided publicly by the authors.
    \item \textbf{Fully synthetic data:} GTA-V \cite{hu2021sail} (in-domain evaluation), DeMamba \cite{chen2024demamba}.
    \item \textbf{In-the-wild DeepFakes:} Publicly available videos from the New York Times DeepFake quiz \cite{nytimesquiz}.
\end{itemize}

\begin{table*}[]
\centering
\caption{Results from the \methodname{} model trained with (1) FF++ \cite{rossler2019faceforensics++} only and (2) FF++ \cite{rossler2019faceforensics++} combined with GTA-V \cite{hu2021sail}. All other results reflect cross-dataset evaluations except for FF++ and GTA-V (when trained). Performance gains are highlighted in {\color[HTML]{009901}{green}}.}
\vspace{-1em}
\resizebox{0.8\textwidth}{!}{
\begin{tabular}{cc|ccccccc}
\hline
\multicolumn{2}{c|}{\textbf{Train}}                            & \multicolumn{7}{c}{\textbf{Test}}                                                                                                                                                                                                                                           \\ \hline
\multicolumn{1}{c|}{\textbf{FF++}} & \textbf{GTA-V} & \multicolumn{1}{c|}{\textbf{Dataset}} & \multicolumn{1}{c|}{\textbf{Accuracy}} & \multicolumn{1}{c|}{\textbf{AUC}} & \multicolumn{1}{c|}{\textbf{Precision@0.5}} & \multicolumn{1}{c|}{\textbf{Recall@0.5}} & \multicolumn{1}{c|}{\textbf{Precs@Rec=0.8}} & \textbf{Rec@Precs=0.8} \\ \hline
\multicolumn{9}{c}{\textit{Face Manipulated Data}} \\ \hline
\multicolumn{1}{c|}{\ding{51}}                     &           & \multicolumn{1}{c|}{FF++}  & \multicolumn{1}{c|}{99.53\%}           & \multicolumn{1}{c|}{99.77\%}       & \multicolumn{1}{c|}{99.94\%}            & \multicolumn{1}{c|}{99.49\%}          & \multicolumn{1}{c|}{99.94\%}                & 99.94\%                \\ 
\multicolumn{1}{c|}{\ding{51}}                     &           & \multicolumn{1}{c|}{CelebDF}          & \multicolumn{1}{c|}{72.61\%}           & \multicolumn{1}{c|}{94.05\%}       & \multicolumn{1}{c|}{96.45\%}            & \multicolumn{1}{c|}{61.22\%}          & \multicolumn{1}{c|}{80.45\%}                & 61.22\%                \\ 
\multicolumn{1}{c|}{\ding{51}}                     &           & \multicolumn{1}{c|}{DeeperForensics}          & \multicolumn{1}{c|}{91.35\%}           & \multicolumn{1}{c|}{100.00\%}       & \multicolumn{1}{c|}{100.00\%}            & \multicolumn{1}{c|}{91.35\%}          & \multicolumn{1}{c|}{100.00\%}                & 91.35\%                \\ 
\multicolumn{1}{c|}{\ding{51}}                     &           & \multicolumn{1}{c|}{DeepfakeTIMIT}          & \multicolumn{1}{c|}{86.90\%}           & \multicolumn{1}{c|}{86.46\%}       & \multicolumn{1}{c|}{83.61\%}            & \multicolumn{1}{c|}{83.97\%}          & \multicolumn{1}{c|}{88.90\%}                & 81.33\%                \\ 
\multicolumn{1}{c|}{\ding{51}}                     &           & \multicolumn{1}{c|}{HifiFace}          & \multicolumn{1}{c|}{63.63\%}           & \multicolumn{1}{c|}{62.47\%}       & \multicolumn{1}{c|}{67.12\%}            & \multicolumn{1}{c|}{63.63\%}          & \multicolumn{1}{c|}{59.30\%}                & 63.63\%                \\ 
\multicolumn{1}{c|}{\ding{51}}                     &           & \multicolumn{1}{c|}{UADFV}            & \multicolumn{1}{c|}{94.12\%}           & \multicolumn{1}{c|}{94.38\%}       & \multicolumn{1}{c|}{95.68\%}            & \multicolumn{1}{c|}{97.11\%}         & \multicolumn{1}{c|}{93.79\%}                & 94.38\%               \\ \arrayrulecolor[gray]{0.85} \hline \arrayrulecolor{black}
\multicolumn{1}{c|}{\ding{51}}                     & \ding{51}           & \multicolumn{1}{c|}{FF++}  & \multicolumn{1}{c|}{99.96\%{\color[HTML]{009901}{(+0.43)}}}           & \multicolumn{1}{c|}{99.89\%{\color[HTML]{009901}{(+0.12)}}}       & \multicolumn{1}{c|}{100.00\%{\color[HTML]{009901}{(+0.06)}}}           & \multicolumn{1}{c|}{99.84\%{\color[HTML]{009901}{(+0.35)}}}          & \multicolumn{1}{c|}{100.00\%{\color[HTML]{009901}{(+0.06)}}}               & 99.96\%{\color[HTML]{009901}{(+0.02)}}               \\ 
\multicolumn{1}{c|}{\ding{51}}                     & \ding{51}           & \multicolumn{1}{c|}{CelebDF}          & \multicolumn{1}{c|}{95.11\%{\color[HTML]{009901}{(+22.50)}}}           & \multicolumn{1}{c|}{94.36\%{\color[HTML]{009901}{(+0.31)}}}       & \multicolumn{1}{c|}{96.82\%{\color[HTML]{009901}{(+0.37)}}}            & \multicolumn{1}{c|}{68.75\%{\color[HTML]{009901}{(+7.53)}}}          & \multicolumn{1}{c|}{96.53\%{\color[HTML]{009901}{(+16.08)}}}                & 68.75\%{\color[HTML]{009901}{(+7.53)}}                \\ 
\multicolumn{1}{c|}{\ding{51}}                     & \ding{51}  & \multicolumn{1}{c|}{DeeperForensics}          & \multicolumn{1}{c|}{99.62\%{\color[HTML]{009901}{(+8.27)}}}           & \multicolumn{1}{c|}{100.00\%{\color[HTML]{009901}{(+0.00)}}}       & \multicolumn{1}{c|}{100.00\%{\color[HTML]{009901}{(+0.00)}}}            & \multicolumn{1}{c|}{99.62\%{\color[HTML]{009901}{(+8.27)}}}          & \multicolumn{1}{c|}{100.00\%{\color[HTML]{009901}{(+0.00)}}}                & 99.63\%{\color[HTML]{009901}{(+8.28)}}                \\ 
\multicolumn{1}{c|}{\ding{51}}                     & \ding{51}  & \multicolumn{1}{c|}{DeepfakeTIMIT}          & \multicolumn{1}{c|}{91.90\%{\color[HTML]{009901}{(+5.00)}}}           & \multicolumn{1}{c|}{91.33\%{\color[HTML]{009901}{(+4.87)}}}       & \multicolumn{1}{c|}{90.45\%{\color[HTML]{009901}{(+6.84)}}}            & \multicolumn{1}{c|}{88.39\%{\color[HTML]{009901}{(+4.42)}}}          & \multicolumn{1}{c|}{100.00\%{\color[HTML]{009901}{(+11.10)}}}                & 91.95\%{\color[HTML]{009901}{(+10.62)}}                \\ 
\multicolumn{1}{c|}{\ding{51}}                     & \ding{51}  & \multicolumn{1}{c|}{HifiFace}          & \multicolumn{1}{c|}{75.62\%{\color[HTML]{009901}{(+11.99)}}}           & \multicolumn{1}{c|}{81.24\%{\color[HTML]{009901}{(+18.77)}}}       & \multicolumn{1}{c|}{79.55\%{\color[HTML]{009901}{(+12.43)}}}            & \multicolumn{1}{c|}{71.71\%{\color[HTML]{009901}{(+8.08)}}}          & \multicolumn{1}{c|}{75.62\%{\color[HTML]{009901}{(+16.32)}}}                & 72.47\%{\color[HTML]{009901}{(+8.84)}}                \\ 
\multicolumn{1}{c|}{\ding{51}}                     & \ding{51}           & \multicolumn{1}{c|}{UADFV}            & \multicolumn{1}{c|}{97.01\%{\color[HTML]{009901}{(+2.89)}}}           & \multicolumn{1}{c|}{94.95\%{\color[HTML]{009901}{(+0.57)}}}       & \multicolumn{1}{c|}{96.89\%{\color[HTML]{009901}{(+1.21)}}}            & \multicolumn{1}{c|}{100.00\%{\color[HTML]{009901}{(+2.89)}}}         & \multicolumn{1}{c|}{94.12\%{\color[HTML]{009901}{(+0.33)}}}                & 100.00\%{\color[HTML]{009901}{(+5.62)}}               \\ \hline
%%%
\multicolumn{9}{c}{\textit{Background Manipulated Data}} \\ \hline
%%%
\multicolumn{1}{c|}{\ding{51}}                     &           & \multicolumn{1}{c|}{AVID}             & \multicolumn{1}{c|}{41.67\%}           & \multicolumn{1}{c|}{33.33\%}      & \multicolumn{1}{c|}{33.33\%}           & \multicolumn{1}{c|}{41.67\%}          & \multicolumn{1}{c|}{41.67\%}                & 33.33\%                \\ \arrayrulecolor[gray]{0.85} \hline \arrayrulecolor{black}
\multicolumn{1}{c|}{\ding{51}}                     & \ding{51}           & \multicolumn{1}{c|}{AVID}             & \multicolumn{1}{c|}{100.00\%{\color[HTML]{009901}{(+58.33)}}}          & \multicolumn{1}{c|}{100.00\%{\color[HTML]{009901}{(+66.67)}}}      & \multicolumn{1}{c|}{100.00\%{\color[HTML]{009901}{(+66.67)}}}           & \multicolumn{1}{c|}{100.00\%{\color[HTML]{009901}{(+58.33)}}}         & \multicolumn{1}{c|}{100.00\%{\color[HTML]{009901}{(+58.33)}}}               & 100.00\%{\color[HTML]{009901}{(+66.67)}}               \\ \hline
%%%
\multicolumn{9}{c}{\textit{Fully Synthetic Data}} \\ \hline
%%%
\multicolumn{1}{c|}{\ding{51}}                     &           & \multicolumn{1}{c|}{GTA-V}            & \multicolumn{1}{c|}{60.16\%}           & \multicolumn{1}{c|}{61.52\%}      & \multicolumn{1}{c|}{60.16\%}           & \multicolumn{1}{c|}{58.73\%}          & \multicolumn{1}{c|}{63.29\%}               & 58.73\%                \\ 
\multicolumn{1}{c|}{\ding{51}}                     &           & \multicolumn{1}{c|}{DeMamba}          & \multicolumn{1}{c|}{61.47\%}           & \multicolumn{1}{c|}{57.38\%}       & \multicolumn{1}{c|}{67.73\%}            & \multicolumn{1}{c|}{33.01\%}          & \multicolumn{1}{c|}{62.15\%}                & 54.16\%                 \\ \arrayrulecolor[gray]{0.85} \hline \arrayrulecolor{black}
\multicolumn{1}{c|}{\ding{51}}                     & \ding{51}           & \multicolumn{1}{c|}{GTA-V}            & \multicolumn{1}{c|}{100.00\%{\color[HTML]{009901}{(+39.84)}}}          & \multicolumn{1}{c|}{100.00\%{\color[HTML]{009901}{(+38.48)}}}      & \multicolumn{1}{c|}{100.00\%{\color[HTML]{009901}{(+39.84)}}}           & \multicolumn{1}{c|}{100.00\%{\color[HTML]{009901}{(+41.27)}}}         & \multicolumn{1}{c|}{100.00\%{\color[HTML]{009901}{(+36.71)}}}               & 100.00\%{\color[HTML]{009901}{(+41.27)}}               \\ 
\multicolumn{1}{c|}{\ding{51}}                     & \ding{51}           & \multicolumn{1}{c|}{DeMamba}          & \multicolumn{1}{c|}{87.12\%{\color[HTML]{009901}{(+25.65)}}}           & \multicolumn{1}{c|}{93.75\%{\color[HTML]{009901}{(+36.67)}}}       & \multicolumn{1}{c|}{92.76\%{\color[HTML]{009901}{(+25.03)}}}            & \multicolumn{1}{c|}{89.60\%{\color[HTML]{009901}{(+56.59)}}}          & \multicolumn{1}{c|}{89.81\%{\color[HTML]{009901}{(+27.66)}}}                & 92.12\%{\color[HTML]{009901}{(+37.96)}}                \\ \hline
%%%
\multicolumn{9}{c}{\textit{In-the-wild DeepFakes}} \\ \hline
%%%
\multicolumn{1}{c|}{\ding{51}}                     &           & \multicolumn{1}{c|}{NYTimes \cite{nytimesquiz}}           & \multicolumn{1}{c|}{50.00\%}           & \multicolumn{1}{c|}{53.74\%}       & \multicolumn{1}{c|}{50.00\%}            & \multicolumn{1}{c|}{25.00\%}          & \multicolumn{1}{c|}{0.00\%}                 & 0.00\%                 \\ \arrayrulecolor[gray]{0.85} \hline \arrayrulecolor{black}
\multicolumn{1}{c|}{\ding{51}}                     & \ding{51}           & \multicolumn{1}{c|}{NYTimes \cite{nytimesquiz}}           & \multicolumn{1}{c|}{80.00\%{\color[HTML]{009901}{(+30.00)}}}           & \multicolumn{1}{c|}{97.42\%{\color[HTML]{009901}{(+43.68)}}}       & \multicolumn{1}{c|}{83.33\%{\color[HTML]{009901}{(+33.33)}}}            & \multicolumn{1}{c|}{83.33\%{\color[HTML]{009901}{(+58.33)}}}          & \multicolumn{1}{c|}{60.00\%{\color[HTML]{009901}{(+60.00)}}}                & 100.00\%{\color[HTML]{009901}{(+100.00)}}                \\ \hline
%%%%%%%%%%%%%%

\end{tabular}
}
\label{tab:synthvd_results}
\vspace{-1.5em}
\end{table*}

%-----------------------------------------------------------
\subsection{Results and Discussion}
\textbf{Evaluation Metrics:} Although most DeepFake detection methods primarily report accuracy, we evaluate \methodname{} using additional metrics: (1) \textit{AUC} (Area Under the Precision-Recall Curve); (2) \textit{Precision@0.5} and (3) \textit{Recall@0.5} at a $0.5$ confidence threshold; (4) \textit{Precs@Rec=0.8} (precision at 0.8 recall); (5) \textit{Rec@Precs=0.8} (recall at 0.8 precision).
% \begin{itemize}
%     \item \textit{AUC (Area Under the Precision-Recall Curve)}: Summarizes the trade-off between precision and recall across different thresholds, providing insight into model performance on imbalanced data.
%     \item \textit{Precision@0.5}: The proportion of true positive predictions among all positive predictions at a confidence threshold of 0.5, reflecting how many of the predicted positives are correct.
%     \item \textit{Recall@0.5}: The proportion of true positive predictions detected out of all actual positive samples at a confidence threshold of 0.5, indicating the model's ability to identify positive cases.
%     \item \textit{Precs@Rec=0.8}: The precision level achieved when the recall is fixed at 0.8, illustrating the model's accuracy in predictions when ensuring 80\% of positive cases are detected.
%     \item \textit{Rec@Precs=0.8}: The recall value when the precision is constrained to be at least 0.8, showing the model's ability to find true positives while maintaining high prediction accuracy.
% \end{itemize}

\noindent
\textbf{Training Details:} \methodname{} is trained using an AdamW optimizer \cite{loshchilov2017decoupled} with an initial learning rate of $0.0001$ and a decay rate of $0.5$ every $1000$ steps. For AD-loss, $\delta_{\text{between}} = 0.5$ and $\delta_{\text{within}}$ is set to $[0.01, -2]$ for binary and $[0.01, -2, 1]$ for fine-grained classification in Sec. \ref{subsec:finegrained} (see supplementary for sensitivity). Loss weights are $\lambda_1 = \lambda_2 = 0.5$. Training uses a batch size of $32$ for $25$ epochs on $8$ TPUv3 chips, with the framework implemented in TensorFlow.

\noindent
\textbf{Quantitative Results:} The results obtained by the \methodname{} model when trained on FF++ \cite{rossler2019faceforensics++} alone, and when trained with both FF++ \cite{rossler2019faceforensics++} and GTA-V \cite{hu2021sail} data and evaluated on the various datasets mentioned in Sec. \ref{subsec:datasets} are shown in Table \ref{tab:synthvd_results}. On the AVID \cite{zhang2024avid} and DeMamba datasets \cite{chen2024demamba}, the model performs poorly when trained with only face manipulated data, but the performance enhances several-fold when the GTA-V \cite{hu2021sail} synthetic data, even though it is not AI-generated, is used in training. 

Interestingly, when evaluating on face-manipulated datasets such as CelebDF \cite{li2020celeb} and DeeperForensics \cite{jiang2020deeperforensics}, we observe a performance boost, particularly for CelebDF \cite{li2020celeb}, when GTA-V \cite{hu2021sail} data is included in the training set. Intuitively, one would expect that training solely on FF++ \cite{rossler2019faceforensics++} would yield similar results since the inference set consists of similar face-manipulated data. This deviation from the expected behavior comes from the contribution of the AD-Loss, as the ablation study (Fig. \ref{fig:ablation}) using only CE-loss compared to both CE and AD-losses reveals that the performance is enhanced with GTA-V \cite{hu2021sail} data in training only in the latter (detailed discussion in Sec. \ref{subsec:ablation}).

To evaluate \methodname{} on in-the-wild DeepFakes, we attempted the recent New York Times quiz \cite{nytimesquiz} (denoted by NYTimes) which provides $10$ videos ($4$ real and $6$ fake) for testing DeepFake detectors. \methodname{} got 8 of those 10 videos correct, even though some of the face-swap videos were indistinguishable from real videos, even by humans. On all the AI-generated videos, \methodname{} made the correct prediction (when trained on FF++ \cite{rossler2019faceforensics++} and GTA-V \cite{hu2021sail}).
% Experiments with different compression factors of the FF++ \cite{rossler2019faceforensics++} dataset are provided in the supplementary.

\begin{table}[h!]
\centering
\caption{\textbf{SOTA Comparison on Face-Manipulated Data:} We compared the performance of \methodname{} with recent DeepFake detectors, in terms of detection accuracy on various face manipulated datasets. \methodname{} outperforms the existing methods. \textbf{Bold} shows the current best results and the previous best and second-best results are highlighted in {\color[HTML]{ff0000}{red}} and {\color[HTML]{0000FF}{blue}} respectively.}\vspace{-0.5em}
\resizebox{0.9\columnwidth}{!}{
\begin{tabular}{c|c|c|c|c}
\hline
\textbf{Method}                                & \textbf{FF++}         & \textbf{CelebDF} & \textbf{DeeperForensics} & \textbf{UADFV}   \\ \hline
TALL \cite{xu2023tall}                         & 98.65\%               & 90.79\%          & {\color[HTML]{ff0000}{99.62\%}}         & -                \\
ISTVT \cite{zhao2023istvt}                     & 99.00\%               & 84.10\%          & 98.60\%                  & -                \\
Concas et al. \cite{concas2024quality}         & 99.49\%               & -                & -                        & -                \\
PUDD  \cite{pellicer2024pudd}                  & -                     & {\color[HTML]{ff0000}{95.10\%}}          & -                        & -                \\
TI2Net \cite{liu2023ti2net}                    & {\color[HTML]{ff0000}{99.95\%}}               & 68.22\%          & 76.08\%                  & -                \\
LRNet \cite{sun2021improving}                  & {\color[HTML]{0000FF}{99.89\%}}               & 53.20\%          & 56.77\%                  & -                \\
Choi et al. \cite{choi2024exploiting}          & \multicolumn{1}{c|}{} & 89.00\%          & {\color[HTML]{0000FF}{99.00\%}}                   & -                \\
Lin et al. \cite{lin2024preserving}            & 98.28\%               & 74.42\%          & -                        & -                \\
Guo et al. \cite{guo2023controllable}          & 99.24\%               & 84.97\%          & -                        & -                \\
Li et al. (Res-152) \cite{li2018exposing}      & -                     & -                & -                        & {\color[HTML]{ff0000}{93.80\%}}           \\
HeadPose \cite{yang2019exposing}               & -                     & -                & -                        & 89.00\%          \\
CViT \cite{wodajo2021deepfake}                 & 93.00\%               & -                & -                        & {\color[HTML]{0000FF}{93.75\%}}           \\
FakeCatcher \cite{ciftci2020fakecatcher}       & 94.65\%               & {\color[HTML]{0000FF}{91.50\%}}           & -                        & -                \\
MesoNet \cite{afchar2018mesonet}               & -                     & -                & -                        & 82.40\%          \\ \arrayrulecolor[gray]{0.85} \hline \arrayrulecolor{black}
\textbf{\methodname{} (Ours)}                  & \textbf{99.96\%}      & \textbf{95.11\%} & \textbf{99.62\%}         & \textbf{97.01\%} \\ \hline
\end{tabular}
}
\label{tab:sota_comparison}
\vspace{-2em}
\end{table}

\begin{figure}
    \centering
    
    \subfloat[Original Frame]{\includegraphics[width=0.3\columnwidth]{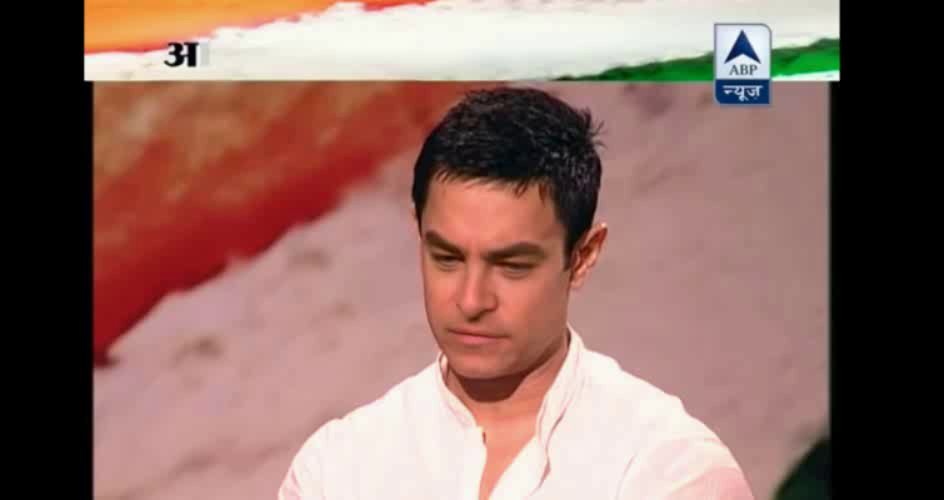}}\hspace{2pt}
    \subfloat[CE Loss only]{\includegraphics[width=0.3\columnwidth]{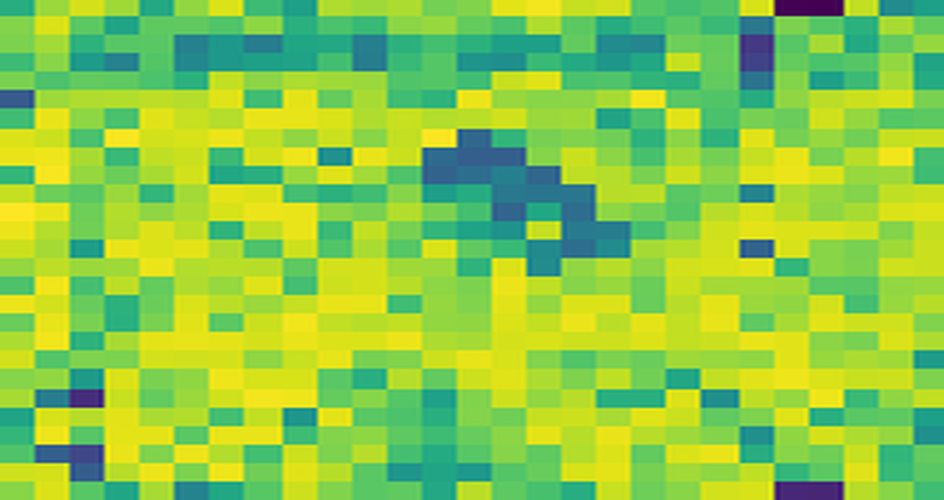}}\hspace{2pt}
    \subfloat[CE+AD Loss]{\includegraphics[width=0.3\columnwidth]{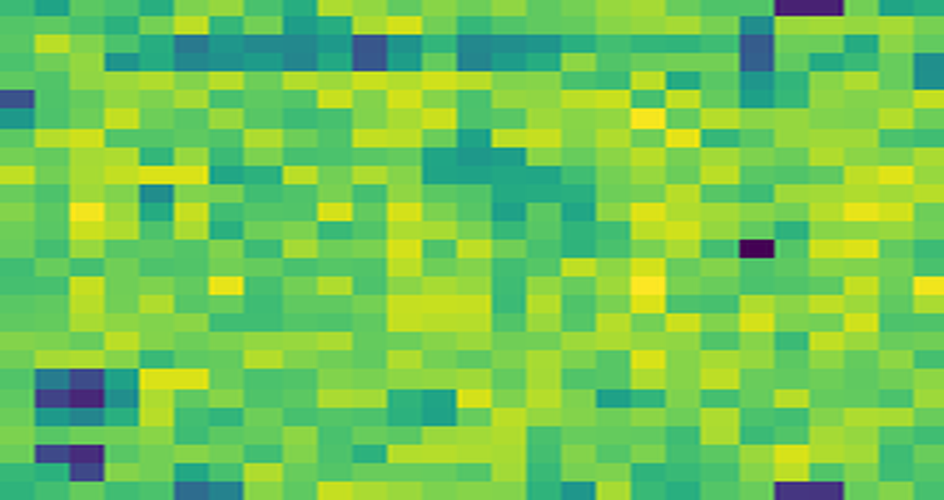}}\\

    \subfloat[Original Frame]{\includegraphics[width=0.3\columnwidth]{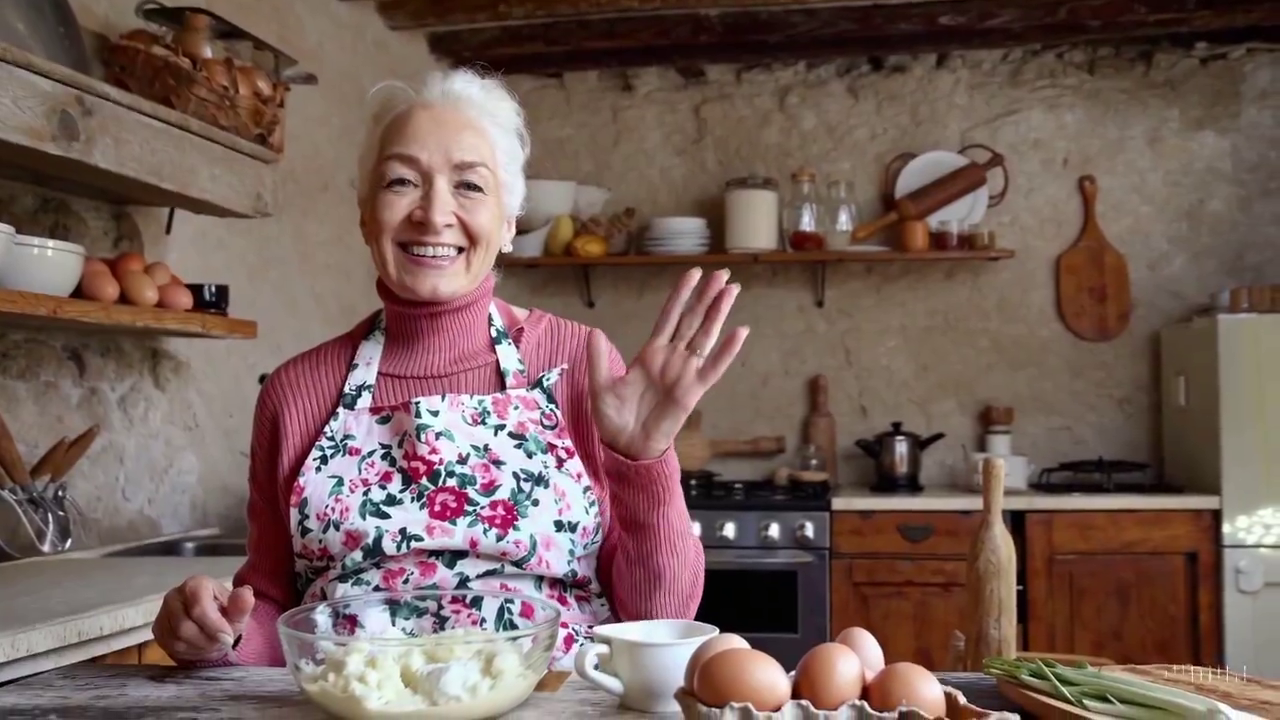}}\hspace{2pt}
    \subfloat[CE Loss only]{\includegraphics[width=0.3\columnwidth]{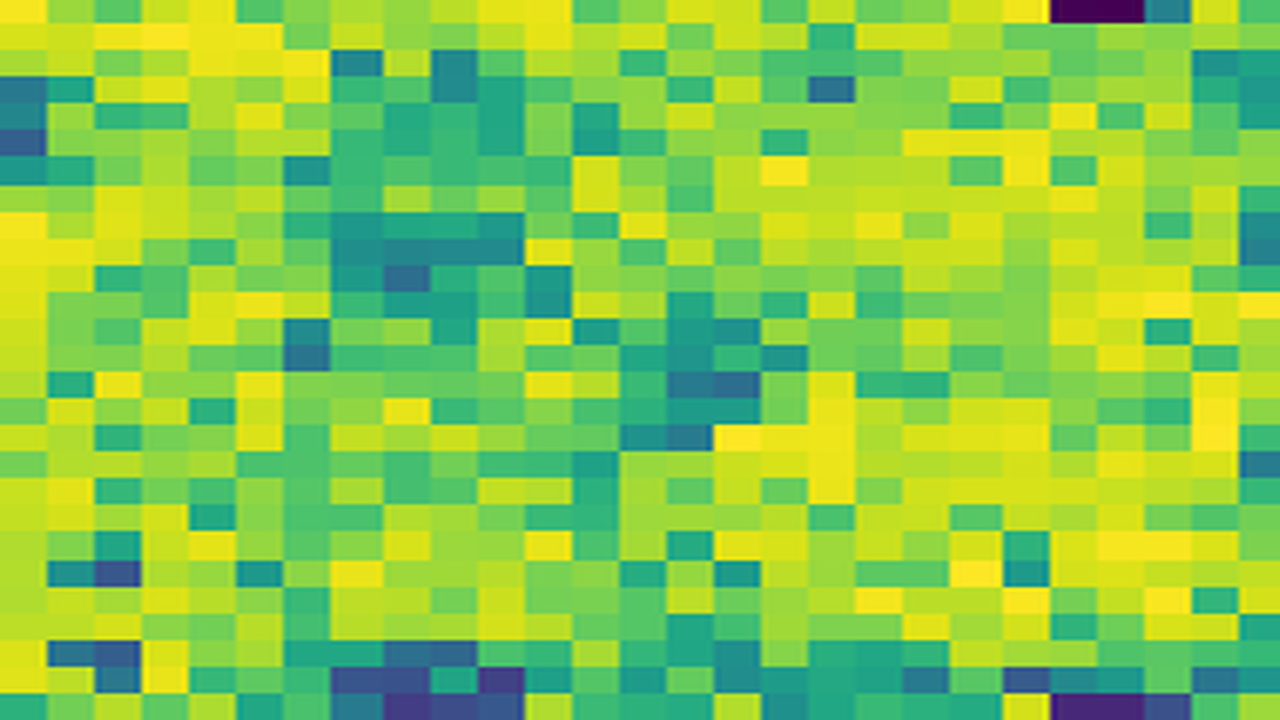}}\hspace{2pt}
    \subfloat[CE+AD Loss]{\includegraphics[width=0.3\columnwidth]{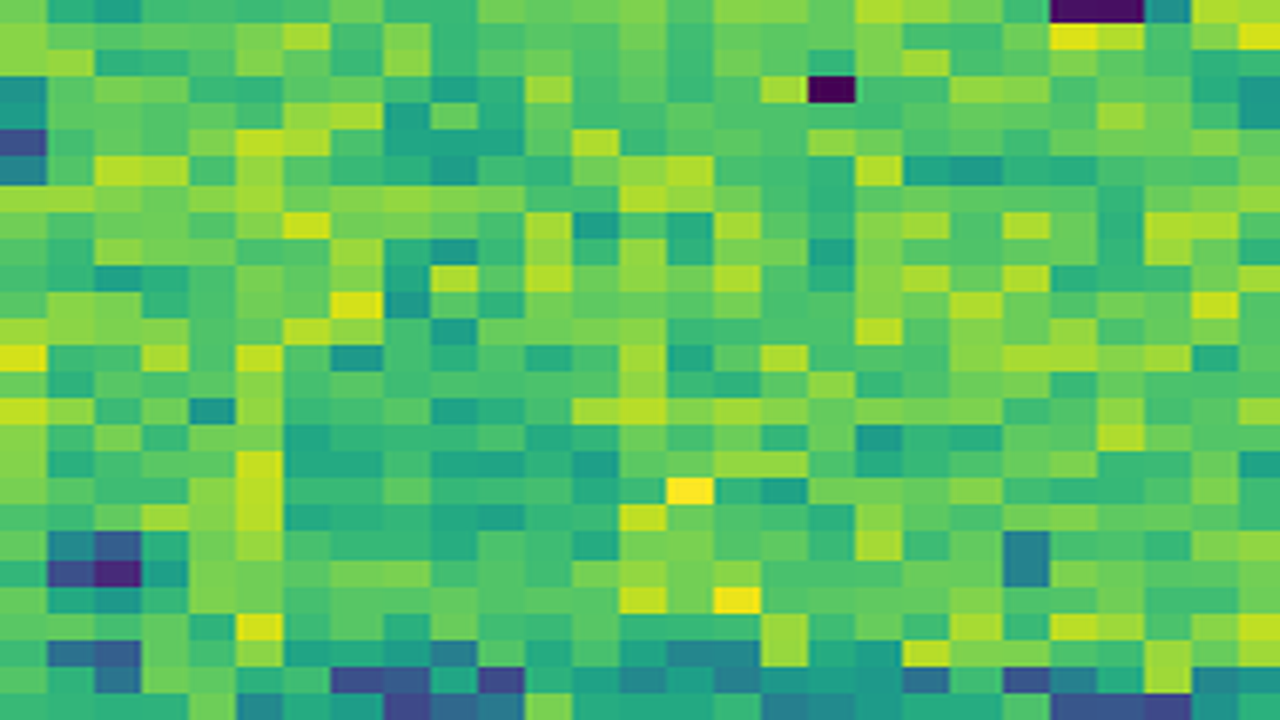}}

    \vspace{-1em}
    \caption{Comparison of attention heatmaps when \methodname{} is trained with only CE loss (second column) versus CE and AD-losses (third column). The CE-loss heatmap predominantly focuses on the face region, whereas the CE+AD loss heatmap demonstrates more distributed attention across the entire frame, as indicated by a broader bluish tone in the heatmaps. Both samples are ``fake" videos-- (a) is taken from Celeb-DF \cite{li2020celeb} and (d) is taken from DeMamba \cite{chen2024demamba} and generated by OpenAI's Sora \cite{brooks2024video} (Best viewed as GIFs provided in the supplementary material). \textcolor{yellow}{Yellow} to \textcolor{blue}{blue} shades show increasing attention in the plot.}
    \vspace{-1.5em}
    \label{fig:heatmaps}
\end{figure}

\noindent
\textbf{Attention Heatmaps:} To analyze the impact of CE and AD-losses on attention distribution, we extracted attention features from the first encoder of the \methodname{} model, comparing models trained with CE loss alone versus that trained with both CE and AD-losses. The resulting heatmaps, shown in Fig. \ref{fig:heatmaps}, reveal that the model trained with only CE loss tends to concentrate primarily on the face region. In contrast, the model trained with the combined CE+AD loss demonstrates a broader attention span across the frame, as evidenced by a lighter, more distributed bluish tone in the heatmaps, indicating increased spatial diversity in the model’s focus. This is why on the example shown in the first row (face-manipulated video) both models give the correct prediction (``fake") with high confidence, but on the row-2 example (video generated from the Sora \cite{brooks2024video} T2V model) the model trained with only CE loss predicted ``real" with a confidence score of $99.25$\% (which is wrong), but the model trained with CE+AD losses predicted ``fake" with a confidence of $100.00$\%, which is correct.

\noindent
\textbf{Comparison on Face Manipulated Data:} \methodname{} is compared with recent state-of-the-art (SOTA) DeepFake detection models in Table \ref{tab:sota_comparison}. The methods we compared against were specifically designed for detecting face manipulations and often crop faces from the videos, during both training and inference of the models. However, \methodname{}, which is designed to detect a diverse range of fake videos, still outperforms these detectors.

\noindent
\textbf{Comparison on Synthetic Data:} We evaluated \methodname{} against state-of-the-art synthetic video detectors on the DeMamba dataset \cite{chen2024demamba} (validation split), with results in Table \ref{tab:sota_synthetic}. While \textit{existing detectors were trained on the DeMamba train split and validated on the DeMamba validation split}, \methodname{} was trained on FF++ \cite{rossler2019faceforensics++} and GTA-V \cite{hu2021sail}. \emph{Despite not being trained on DeMamba, \methodname{}'s average performance (including real videos) surpassed current SOTA detectors, with competitive generator-wise results.}

Thus, \methodname{} reliably detects a diverse range of fake videos (even in cross-domain settings), including T2V/I2V videos and background manipulations, while consistently maintaining state-of-the-art performance on traditional face-manipulated data. \methodname{} eliminates the need to have separate DeepFake and T2V/I2V video detector models by by handling both tasks within a single trained model.

\begin{table*}[]
\centering
\caption{\textbf{SOTA Comparison on Synthetic Data:} On the DeMamba dataset (validation split), we compare the performance of \underline{\methodname{}, which was \textit{NOT} trained on DeMamba} train split, against \underline{state-of-the-art detectors \textit{which were trained on DeMamba train split}} (results taken from Chen et al. \cite{chen2024demamba}). We report the results ($P$ = Precision and $R$ = Recall) on the individual T2V/I2V generators and the average performance across the entire validation set ($Avg$, which also includes real videos). Although the direct comparison is unfair against \methodname{} which was trained with FF++ \cite{rossler2019faceforensics++} and GTA-V \cite{hu2021sail}, our method still outperforms these synthetic video detectors. \textbf{Bold} shows the current best results and the previous best and second-best results are highlighted in {\color[HTML]{ff0000}{red}} and {\color[HTML]{0000FF}{blue}} respectively.}\vspace{-1em}
\resizebox{0.95\textwidth}{!}{
\begin{tabular}{c|c|c|c|c|c|c|c|c|c|c|c|c}
\hline
\textbf{Method}                                                  & \textbf{Metrics} & \textbf{Sora} \cite{brooks2024video}         & \begin{tabular}[c]{@{}c@{}}\textbf{Morph}\\ \textbf{Studio} \cite{morphstudio}\end{tabular} & \begin{tabular}[c]{@{}c@{}}\textbf{Runway ML}\\ \textbf{(Gen2)} \cite{runwayresearch}\end{tabular}         & \textbf{HotShot} \cite{Mullan_Hotshot-XL_2023}      & \textbf{Lavie \cite{wang2023lavie}}        & \textbf{Show-1} \cite{zhang2024show}       & \begin{tabular}[c]{@{}c@{}}\textbf{Moon}\\ \textbf{Valley} \cite{moonvalley2022}\end{tabular} & \textbf{Crafter} \cite{chen2023videocrafter1}      & \begin{tabular}[c]{@{}c@{}}\textbf{Model}\\ \textbf{Scope} \cite{wang2023modelscope}\end{tabular} & \begin{tabular}[c]{@{}c@{}}\textbf{Wild}\\ \textbf{Scrape} \cite{chen2024demamba}\end{tabular} & \textbf{Avg}                   \\ \hline
                                                                 & P                & 71.15\%                        & 96.89\%                                                         & 98.51\%                                                             & 79.38\%                        & 84.59\%                        & 79.38\%                        & 98.79\%                                                        & 99.02\%                         & 92.70\%                                                        & 76.47\%                                                        & 87.91\%                        \\  
\multirow{-2}{*}{TALL \cite{xu2023tall}}                                           & R                & {\color[HTML]{0000FF} 91.07\%} & 98.28\%                                                         & 97.83\%                                                             & {\color[HTML]{0000FF} 83.00\%} & 76.57\%                        & 79.57\%                        & 99.52\%                                                        & 98.93\%                         & 94.14\%                                                        & 66.31\%                                                        & {\color[HTML]{FF0000} 88.52\%} \\ \arrayrulecolor[gray]{0.85} \hline \arrayrulecolor{black}
                                                                 & P                & 68.27\%                        & 99.89\%                                                         & 99.67\%                                                             & {\color[HTML]{FF0000} 89.35\%} & 57.00\%                        & 36.57\%                        & 99.52\%                                                        & 99.71\%                         & 93.80\%                                                        & 88.41\%                                                        & 88.73\%                        \\  
\multirow{-2}{*}{F3Net \cite{qian2020thinking}}                                          & R                & 83.93\%                        & 99.71\%                                                         & 98.62\%                                                             & 77.57\%                        & 85.24\%                        & 63.17\%                        & {\color[HTML]{0000FF} 99.58\%}                                 & 99.89\%                         & 89.43\%                                                        & 76.78\%                                                        & 81.88\%                        \\ \arrayrulecolor[gray]{0.85} \hline \arrayrulecolor{black}
                                                                 & P                & {\color[HTML]{FF0000} 91.07\%} & 99.57\%                                                         & 99.49\%                                                             & 24.29\%                        & 89.64\%                        & 57.71\%                        & 97.12\%                                                        & 99.86\%                         & 94.29\%                                                        & 87.80\%                                                        & 82.45\%                        \\  
\multirow{-2}{*}{NPR \cite{tan2024rethinking}}                                            & R                & {\color[HTML]{FF0000} 91.07\%} & 99.57\%                                                         & \textbf{99.49\%}                                                    & 24.29\%                        & 89.64\%                        & 57.71\%                        & 97.12\%                                                        & 99.86\%                         & 94.29\%                                                        & {\color[HTML]{FF0000} 87.80\%}                                 & 84.08\%                        \\ \arrayrulecolor[gray]{0.85} \hline \arrayrulecolor{black}
                                                                 & P                & 57.21\%                        & 99.08\%                                                         & 99.32\%                                                             & {\color[HTML]{0000FF} 86.19\%} & 82.24\%                        & 70.43\%                        & 99.25\%                                                        & 98.96\%                         & {\color[HTML]{FF0000} 97.18\%}                                 & 81.32\%                                                        & 87.12\%                        \\  
\multirow{-2}{*}{STIL \cite{gu2021spatiotemporal}}                                           & R                & 67.86\%                        & 96.00\%                                                         & 98.41\%                                                             & {\color[HTML]{FF0000} 96.14\%} & 77.14\%                        & 80.43\%                        & 97.44\%                                                        & 96.93\%                         & {\color[HTML]{FF0000} 96.29\%}                                 & 68.36\%                                                        & 82.22\%                        \\ \arrayrulecolor[gray]{0.85} \hline \arrayrulecolor{black}
                                                                 & P                & 83.21\%                        & {\color[HTML]{FF0000} 99.99\%}                                  & {\color[HTML]{0000FF} 99.67\%}                                      & 50.84\%                        & {\color[HTML]{FF0000} 99.20\%} & \textbf{99.27\%}               & {\color[HTML]{FF0000} 99.76\%}                                 & {\color[HTML]{0000FF} 99.99\%}  & 91.83\%                                                        & {\color[HTML]{0000FF} 91.77\%}                                 & {\color[HTML]{0000FF} 91.55\%} \\  
\multirow{-2}{*}{MINTIME-CLIP-B \cite{chen2024demamba}}                                 & R                & 89.29\%                        & {\color[HTML]{FF0000} 100.00\%}                                 & {\color[HTML]{FF0000} 98.99\%}                                      & 26.43\%                        & {\color[HTML]{0000FF} 96.79\%} & {\color[HTML]{FF0000} 98.14\%} & {\color[HTML]{FF0000} 99.84\%}                                 & {\color[HTML]{0000FF} 100.00\%} & 84.29\%                                                        & 82.38\%                                                        & {\color[HTML]{0000FF} 87.62\%} \\ \arrayrulecolor[gray]{0.85} \hline \arrayrulecolor{black}
                                                                 & P                & \textbf{91.79\%}               & {\color[HTML]{0000FF} 99.99\%}                                  & {\color[HTML]{FF0000} 99.79\%}                                      & 45.94\%                        & \textbf{99.76\%}               & {\color[HTML]{0000FF} 97.80\%} & \textbf{99.99\%}                                               & {\color[HTML]{FF0000} 99.99\%}  & 94.69\%                                                        & {\color[HTML]{FF0000} 92.32\%}                                 & {\color[HTML]{FF0000} 92.21\%} \\  
\multirow{-2}{*}{FTCN-CLIP-B \cite{chen2024demamba}}                                    & R                & 87.50\%                        & {\color[HTML]{0000FF} 100.00\%}                                 & {\color[HTML]{0000FF} 98.91\%}                                      & 17.71\%                        & {\color[HTML]{FF0000} 97.71\%} & {\color[HTML]{0000FF} 91.86\%} & \textbf{100.00\%}                                              & {\color[HTML]{FF0000} 100.00\%} & 85.29\%                                                        & 82.83\%                                                        & 86.18\%                        \\ \arrayrulecolor[gray]{0.85} \hline \arrayrulecolor{black}
                                                                 & P                & 67.80\%                        & 43.56\%                                                         & 70.88\%                                                             & 29.97\%                        & 52.97\%                        & 35.36\%                        & 55.52\%                                                        & 66.03\%                         & 44.23\%                                                        & 42.99\%                                                        & 44.83\%                        \\  
\multirow{-2}{*}{CLIP-B-PT \cite{chen2024demamba}}                                      & R                & 85.71\%                        & 82.43\%                                                         & 90.36\%                                                             & 71.00\%                        & 79.29\%                        & 75.43\%                        & 89.62\%                                                        & 86.29\%                         & 82.14\%                                                        & 75.16\%                                                        & 81.74\%                        \\ \arrayrulecolor[gray]{0.85} \hline \arrayrulecolor{black}
                                                                 & P                & 25.87\%                        & 95.14\%                                                         & 96.23\%                                                             & 73.43\%                        & 83.31\%                        & 75.49\%                        & 90.17\%                                                        & 95.06\%                         & {\color[HTML]{0000FF} 95.05\%}                                 & 69.95\%                                                        & 79.97\%                        \\  
\multirow{-2}{*}{DeMamba-CLIP-PT \cite{chen2024demamba}}                                & R                & 58.93\%                        & 96.43\%                                                         & 93.12\%                                                             & 68.00\%                        & 69.36\%                        & 69.00\%                        & 89.14\%                                                        & 91.86\%                         & {\color[HTML]{0000FF} 96.14\%}                                 & 56.59\%                                                        & 78.86\%                        \\ \arrayrulecolor[gray]{0.85} \hline \arrayrulecolor{black}
                                                                 & P                & 16.39\%                        & 72.16\%                                                         & 87.77\%                                                             & 39.86\%                        & 65.57\%                        & 54.26\%                        & 75.23\%                                                        & 84.80\%                         & 61.60\%                                                        & 55.28\%                                                        & 61.29\%                        \\  
\multirow{-2}{*}{XCLIP-B-PT \cite{chen2024demamba}}                                     & R                & 81.34\%                        & 82.15\%                                                         & 83.35\%                                                             & 80.98\%                        & 81.82\%                        & 81.55\%                        & 82.14\%                                                        & 82.98\%                         & 81.93\%                                                        & 81.10\%                                                        & 81.93\%                        \\ \arrayrulecolor[gray]{0.85} \hline \arrayrulecolor{black}
                                                                 & P                & 18.26\%                        & 93.50\%                                                         & 94.72\%                                                             & 69.94\%                        & 78.08\%                        & 71.50\%                        & 83.95\%                                                        & 92.23\%                         & 93.54\%                                                        & 68.10\%                                                        & 76.38\%                        \\  
\multirow{-2}{*}{DeMamba-XCLIP-PT \cite{chen2024demamba}}                               & R                & 66.07\%                        & 95.86\%                                                         & 94.64\%                                                             & 77.86\%                        & 75.36\%                        & 80.29\%                        & 90.89\%                                                        & 92.50\%                         & 96.00\%                                                        & 66.41\%                                                        & 83.59\%                        \\ \arrayrulecolor[gray]{0.85} \hline \arrayrulecolor{black}
                                                                 & P                & 64.42\%                        & 99.73\%                                                         & 96.78\%                                                             & 70.98\%                        & {\color[HTML]{0000FF} 90.35\%} & 77.28\%                        & 97.34\%                                                        & 99.84\%                         & 82.01\%                                                        & 88.97\%                                                        & 86.77\%                        \\  
\multirow{-2}{*}{XCLIP-B-FT \cite{chen2024demamba}}                                     & R                & 82.14\%                        & 99.57\%                                                         & 93.62\%                                                             & 61.29\%                        & 79.36\%                        & 69.71\%                        & 97.92\%                                                        & 99.79\%                         & 77.14\%                                                        & {\color[HTML]{0000FF} 83.59\%}                                 & 84.41\%                        \\ \hline
                                                                 & P                & {\color[HTML]{0000FF} 88.57\%} & \textbf{100.00\%}                                               & \textbf{100.00\%}                                                   & \textbf{90.16\%}               & 89.91\%                        & {\color[HTML]{FF0000} 98.34\%} & {\color[HTML]{0000FF} 99.52\%}                                 & \textbf{100.00\%}               & \textbf{98.96\%}                                               & \textbf{92.56\%}                                               & \textbf{92.76\%}               \\  
\multirow{-2}{*}{\textbf{\methodname{} (Ours)}} & R                & \textbf{92.11\%}               & \textbf{100.00\%}                                               & 94.62\%                                                             & \textbf{96.93\%}               & \textbf{98.12\%}               & \textbf{99.86\%}               & 98.69\%                                                        & \textbf{100.00\%}               & \textbf{96.29\%}                                               & \textbf{89.89\%}                                               & \textbf{89.60\%}               \\ \hline
\end{tabular}
}
\label{tab:sota_synthetic}
\vspace{-1.5em}
\end{table*}

\subsection{Finegrained Evaluation}\label{subsec:finegrained}
% In the context of fake media detection, when we try to flag videos which have been manipulated either partially or fully using AI, it also becomes pertinent to look at finegrained classification- where we define three distinct classes as: (1) real videos (2) partially manipulated videos and (3) fully synthetic videos. When trying to prohibit the propagation of misinformation, it might be enough to do real vs. fake binary classification. However, such a finegrained classification allows a little more explainability in an otherwise blackbox model.
In fake media detection, flagging videos partially or fully manipulated by AI also necessitates fine-grained classification into three classes: (1) real videos, (2) partially manipulated videos, and (3) fully synthetic videos. While binary real vs. fake classification may suffice to curb misinformation, fine-grained classification offers greater explainability in otherwise black-box models.

\begin{table}[]
\centering
% \vspace{-1em}
\caption{Results obtained by the \methodname{} model on \textbf{finegrained classes}: detecting whether a video is real, partially manipulated or fully AI-generated. Performance gains are mentioned in {\color[HTML]{009901}{green}}.}\vspace{-0.5em}
\resizebox{0.8\columnwidth}{!}{
\begin{tabular}{cc|cc}
\hline
\multicolumn{2}{c|}{\textbf{Train}}                            & \multicolumn{2}{c}{\textbf{Test}}                                   \\ \hline
\multicolumn{1}{c|}{\textbf{FF++}} & \textbf{GTA-V} & \multicolumn{1}{c|}{\textbf{Dataset}} & \textbf{Accuracy}            \\ \hline
\multicolumn{1}{c|}{\ding{51}}                     &                & \multicolumn{1}{c|}{FF++}  & 96.46\%                      \\
\multicolumn{1}{c|}{\ding{51}}                     &                & \multicolumn{1}{c|}{CelebDF}          & 60.40\% 
       \\
\multicolumn{1}{c|}{\ding{51}}                     &                & \multicolumn{1}{c|}{DeeperForensics}  & 71.04\%                      \\
\multicolumn{1}{c|}{\ding{51}}                     &                & \multicolumn{1}{c|}{DeepFakeTIMIT}    & 80.68\%                      \\
\multicolumn{1}{c|}{\ding{51}}                     &                & \multicolumn{1}{c|}{HifiFace}         & 43.27\%                      \\
\multicolumn{1}{c|}{\ding{51}}                     &                & \multicolumn{1}{c|}{UADFV}            & 90.37\% 
       \\
\multicolumn{1}{c|}{\ding{51}}                     &                & \multicolumn{1}{c|}{AVID}             & 50.00\%                      \\
\multicolumn{1}{c|}{\ding{51}}                     &                & \multicolumn{1}{c|}{GTA-V}            & 0.00\%                       \\
\multicolumn{1}{c|}{\ding{51}}                     &                & \multicolumn{1}{c|}{DeMamba}          & 38.29\% 
       \\ \hline
\multicolumn{1}{c|}{\ding{51}}                     & \ding{51}           & \multicolumn{1}{c|}{FF++}        & 97.70\%{\color[HTML]{009901}{(+1.24)}}                      \\
\multicolumn{1}{c|}{\ding{51}}                     & \ding{51}           & \multicolumn{1}{c|}{CelebDF}          & 70.17\%{\color[HTML]{009901}{(+9.77)}} 
       \\
\multicolumn{1}{c|}{\ding{51}}                     & \ding{51}           & \multicolumn{1}{c|}{DeeperForensics}  & 80.59\%{\color[HTML]{009901}{(+9.55)}}                      \\
\multicolumn{1}{c|}{\ding{51}}                     & \ding{51}           & \multicolumn{1}{c|}{DeepFakeTIMIT}    & 81.04\%{\color[HTML]{009901}{(+0.36)}}                      \\
\multicolumn{1}{c|}{\ding{51}}                     & \ding{51}           & \multicolumn{1}{c|}{HifiFace}         & 64.20\%{\color[HTML]{009901}{(+20.93)}}                      \\
\multicolumn{1}{c|}{\ding{51}}                     & \ding{51}           & \multicolumn{1}{c|}{UADFV}            & 92.61\%{\color[HTML]{009901}{(+1.24)}} 
       \\
\multicolumn{1}{c|}{\ding{51}}                     & \ding{51}           & \multicolumn{1}{c|}{AVID}             & 62.50\%{\color[HTML]{009901}{(+12.50)}}                      \\
\multicolumn{1}{c|}{\ding{51}}                     & \ding{51}           & \multicolumn{1}{c|}{GTA-V}            & 100.00\%{\color[HTML]{009901}{(+100.00)}}                     \\
\multicolumn{1}{c|}{\ding{51}}                     & \ding{51}           & \multicolumn{1}{c|}{DeMamba}          & 65.84\%{\color[HTML]{009901}{(+27.55)}} 
       \\ \hline
\end{tabular}
}
\label{tab:finegrained_results}
\vspace{-2em}
\end{table}

Using this defined convention of finegrained classes, all the face manipulated datasets (CelebDF \cite{li2020celeb}, DeeperForensics \cite{jiang2020deeperforensics}, etc.) and the videos from the AVID \cite{zhang2024avid} model (background manipulations) fall under class-2, and the videos from the GTA-V \cite{hu2021sail} and the DeMamba \cite{chen2024demamba} datasets fall in class-3. In this setup, only the final classification layer of the \methodname{} transformer architecture is changed to $3$ neurons with softmax activation. The results obtained are presented in Table \ref{tab:finegrained_results}. As expected, when the model is trained only with FF++ \cite{rossler2019faceforensics++}, the performance on fully synthetic data is poor (since the model has not seen any synthetic videos). When the GTA-V \cite{hu2021sail} data is also used in training, the performance is much higher.

%-----------------------------------------------------------
\subsection{Ablation Study}\label{subsec:ablation}

\begin{figure}
    \centering
    \subfloat[\methodname{} trained on FF++ \cite{rossler2019faceforensics++} only]{\includegraphics[width=0.9\columnwidth]{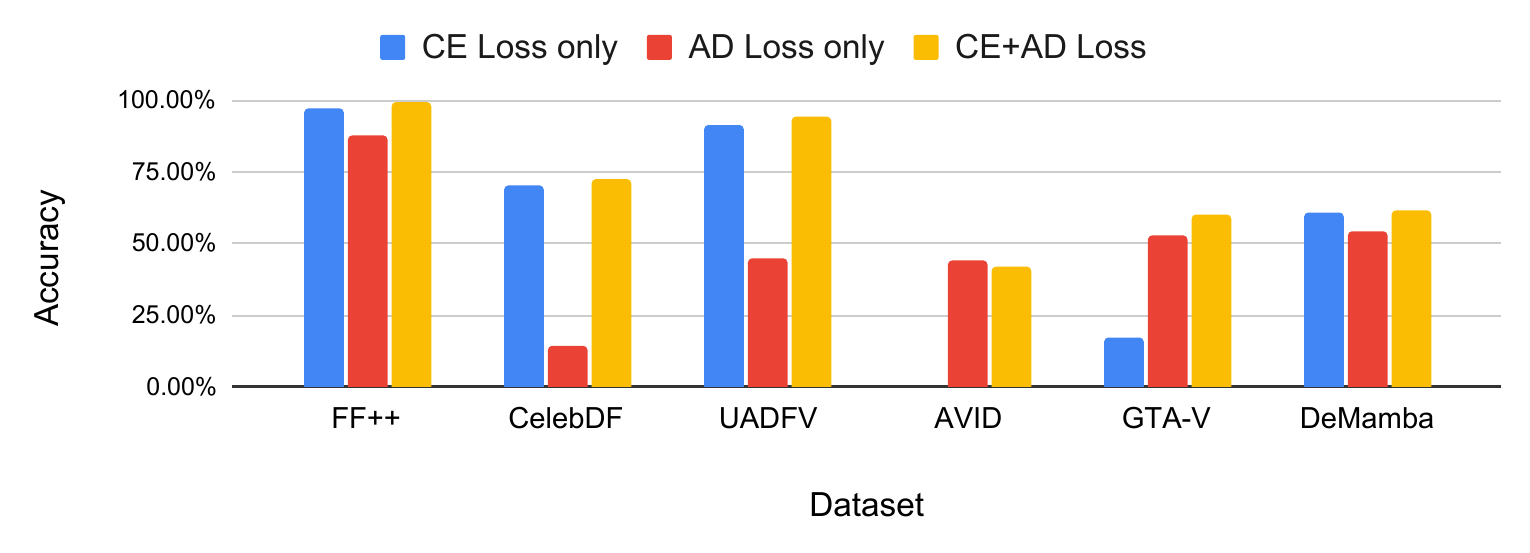}\vspace{-0.5em}}\\
    \subfloat[\methodname{} trained on FF++ \cite{rossler2019faceforensics++} and GTA-V \cite{hu2021sail}]{\includegraphics[width=0.9\columnwidth]{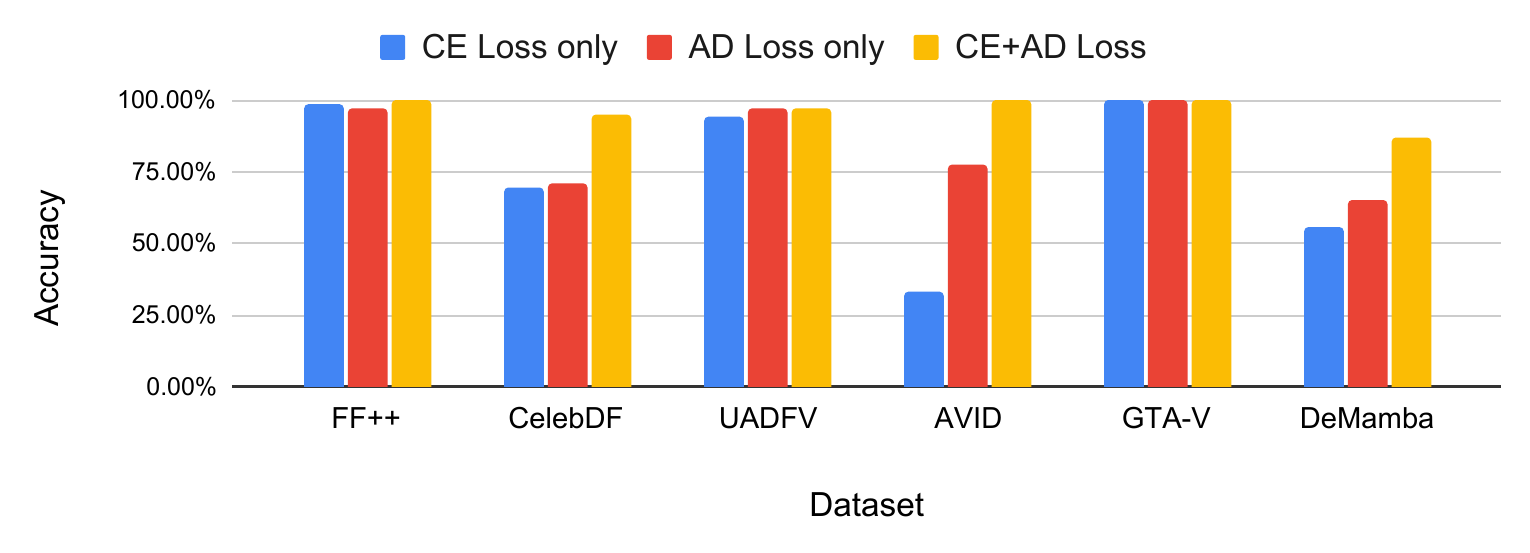}\vspace{-0.5em}}\vspace{-1em}
    \caption{\textbf{Ablation Results} to show the effect of changing the loss functions used to train the \methodname{} model. The combination of the cross-entropy (CE) and Attention Diversity (AD) losses always performs better, and the contribution from the AD-loss component increases significantly when using fully synthetic data for training.}
    \label{fig:ablation}
    \vspace{-1.5em}
\end{figure}

We conducted an analysis to assess the contribution of individual loss functions to the training of the \methodname{} model. The outcomes of this ablation study are presented in Fig. \ref{fig:ablation}. Specifically, we evaluated the impact of the AD-loss in two scenarios: (1) when \methodname{} was trained solely on the FF++ dataset \cite{rossler2019faceforensics++} (Fig. \ref{fig:ablation}(a)), and (2) when \methodname{} was trained using a combination of the FF++ \cite{rossler2019faceforensics++} and the GTA-V \cite{hu2021sail} datasets (Fig. \ref{fig:ablation}(b)). The combination of the two losses always performed better than the individual losses.

\noindent
\textbf{Effect on Synthetic Data:} \methodname{} performed significantly better on the fully sythetic datasets GTA-V \cite{hu2021sail} and DeMamba \cite{chen2024demamba} when the model was training includes GTA-V \cite{hu2021sail} data. However, for the DeMamba dataset (cross-data evaluation), the performance with only CE-loss is similar regardless of the training data. It is the AD-loss component that provides the accuracy boost in Fig \ref{fig:ablation}(b).

\noindent
\textbf{Effect on Background-Manipulated Data}: In Fig. \ref{fig:ablation}(a), we observe that when \methodname{} was trained solely on FF++ \cite{rossler2019faceforensics++} using only the CE-loss, its performance on the AVID dataset \cite{zhang2024avid} was $0.00$\%. This outcome can be attributed to the fact that the fake data in FF++ \cite{rossler2019faceforensics++} contains exclusively face-manipulated videos, causing the \methodname{} transformer's attention heads to focus predominantly on the facial regions. As AVID \cite{zhang2024avid} comprises only background-manipulated videos, \methodname{} failed to detect any of the fakes under such conditions. However, when the AD-loss was introduced, the attention heads of \methodname{} were encouraged to diversify their spatial focus, which significantly improved performance. Incorporating synthetic data into training, in Fig. \ref{fig:ablation}(b), further enhanced this effect, achieving a $100$\% accuracy on the AVID \cite{zhang2024avid} dataset when both CE-loss and AD-loss were applied together.

\noindent
\textbf{Effect on Face-Manipulated Data}: For the evaluation on face-manipulated datasets such as CelebDF \cite{li2020celeb} and UADFV \cite{yang2019exposing}, one might expect comparable results between training solely on FF++ \cite{rossler2019faceforensics++} and incorporating GTA-V \cite{hu2021sail} due to their perceived irrelevance in evaluating face-centric manipulations. However, as shown in Fig. \ref{fig:ablation}(b), training with fully synthetic data leads to a substantial performance improvement. Notably, in both Fig. \ref{fig:ablation}(a) and (b), using only CE loss yields similar results, but the inclusion of the AD-loss component significantly enhances overall detection accuracy. The diversity in the training data makes the \methodname{} model learn more discriminative attention map features through the AD-loss component, which in turn helps even in detecting face-manipulated videos.

\begin{figure}
    \centering
    \includegraphics[width=0.8\columnwidth]{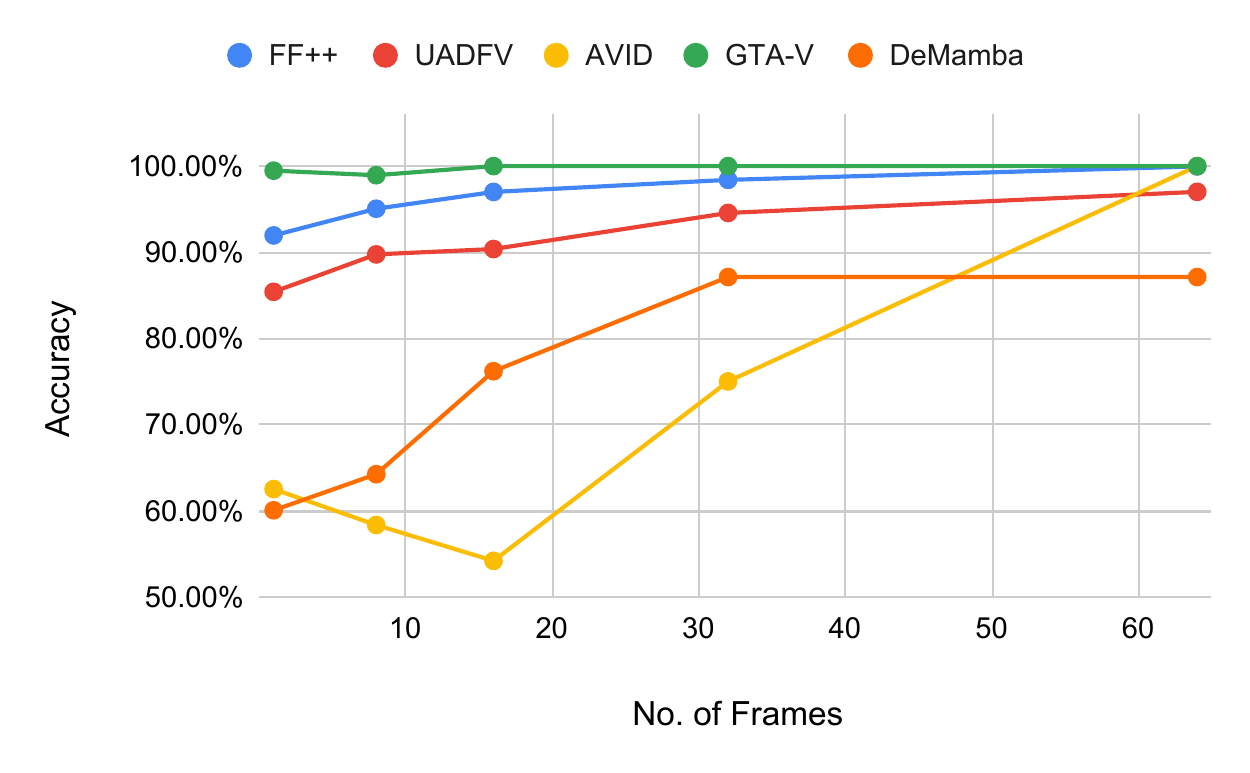}\vspace{-1.5em}
    \caption{\textbf{No. of frames vs. Performance:} Performance analysis of \methodname{} based on the number of frames sampled per video segment. The results illustrate that as the number of frames increases from 1 to 64 (context window), the detection accuracy improves, showcasing \methodname{}'s ability to effectively capture temporal inconsistencies in fake videos.}
    \label{fig:frames_vs_performance}
    \vspace{-1.5em}
\end{figure}

\noindent
\textbf{Frames vs. Performance:} For inference with \methodname{}, we evaluated the impact of temporal context by varying the number of frames sampled from video segments to $\{1, 8, 16, 32, 64\}$, with 64 frames corresponding to our model's context window. As shown in Fig. \ref{fig:frames_vs_performance}, performance improves as more frames are included, demonstrating that \methodname{} effectively captures temporal discontinuities in fake videos. This highlights the model's ability to leverage temporal cues for enhanced detection accuracy.

\begin{figure}
    \centering
    \includegraphics[width=0.8\columnwidth]{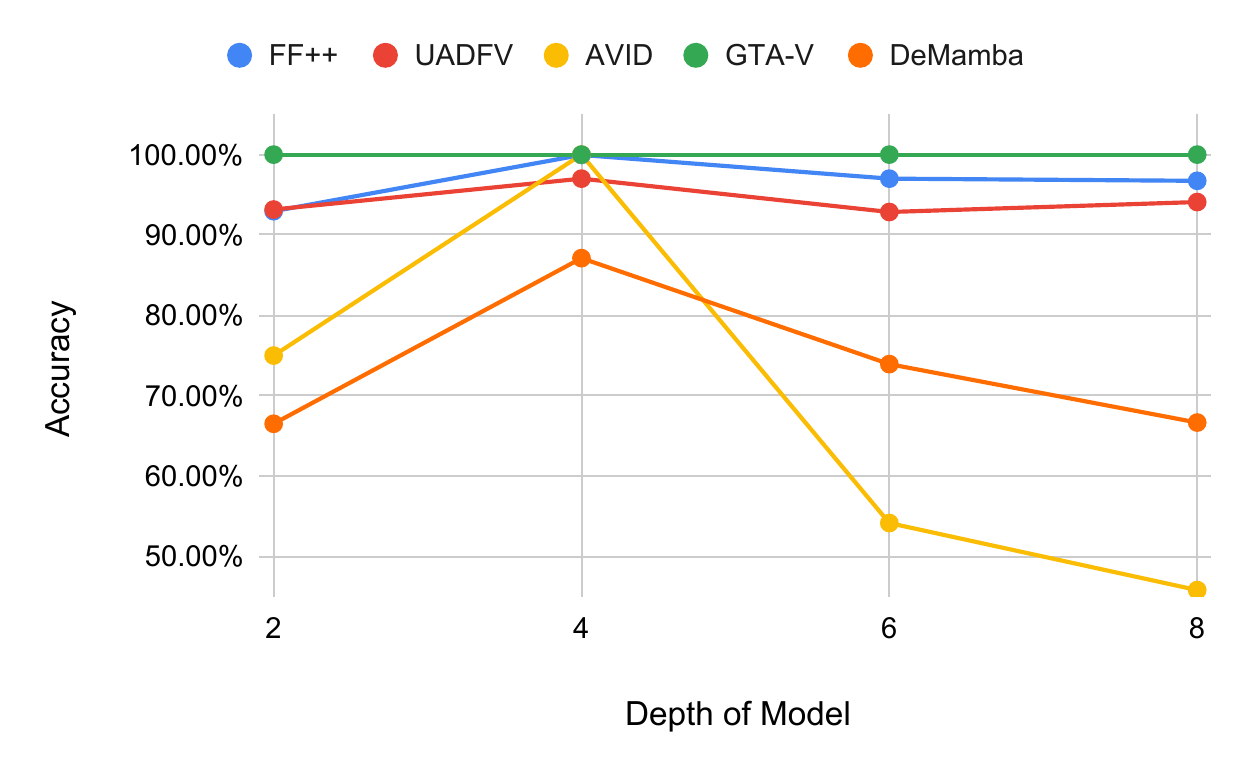}\vspace{-1em}
    \caption{\textbf{Transformer Depth Evaluation:} Performance comparison of \methodname{} with varying the number of encoder blocks (depth). \methodname{} performs optimally in cross-domain settings when the depth is 4. Greater depths overfit to the training domains (FF++ \cite{rossler2019faceforensics++} and GTA-V \cite{hu2021sail}), while a depth of 2 is insufficient to capture the complexity of the data.}
    \vspace{-1.5em}
    \label{fig:depth_ablation}
\end{figure}

\noindent
\textbf{Transformer Depth Ablation:} We evaluate the impact of varying the depth of the transformer architecture in \methodname{} by adjusting the number of encoder blocks, with results shown in Fig. \ref{fig:depth_ablation}. We tested depths of $\{2, 4, 6, 8\}$ and observed that the model achieved the best performance with $4$ encoder blocks, which is the depth used in our final architecture. This suggests that a moderate depth strikes the optimal balance between model complexity and performance, especially in cross-dataset settings, providing robust detection capabilities without overfitting.  
\section{Conclusion} \label{sec:conclusion}
% Traditional DeepFake detectors are increasingly becoming outdated as they primarily focus on detecting face-manipulated content, which leaves them vulnerable to newer, more sophisticated methods of forgery. The rapid advancement of text-to-video or image-to-video generative models, and models that can manipulate the background necessitate a model capable of more comprehensive detection. To address these challenges, we proposed \methodname{}, a novel approach that leverages the fully synthetic data GTA-V \cite{hu2021sail} for training. While the synthetic data used is not AI-generated, our results demonstrate that it effectively enables \methodname{} to detect AI-generated content and background manipulations with high reliability.

Traditional DeepFake detectors, focused primarily on face-manipulated content, struggle against newer, more sophisticated forgery methods involving T2V, I2V, and background manipulations. To address this, we proposed \methodname{}, leveraging the fully synthetic GTA-V \cite{hu2021sail} dataset for training. While this synthetic data is not AI-generated, our results show it effectively enhances \methodname{}'s ability to detect AI-generated content and background manipulations.

A key innovation in our approach is the Attention-Diversity (AD) loss, which encourages the model's attention mechanism to explore diverse spatial regions, improving its detection of manipulated areas beyond the face. Consequently, \methodname{} not only excels at detecting AI-generated and background-altered videos but also shows a marked improvement in detecting traditional face-manipulated content. The comprehensive performance gains highlighted through our ablation studies underscore the robustness of \methodname{}, making it a valuable tool in the evolving landscape of synthetic video detection.

% A key innovation in our approach is the Attention-Diversity (AD) loss, which encourages the model's attention mechanism to explore diverse spatial regions, improving its detection of manipulated areas beyond the face. Consequently, \methodname{} excels in identifying both AI-generated, background-altered, and traditional face-manipulated content, demonstrating strong performance gains and robustness as validated through our ablation studies.
{
    \small
    \bibliographystyle{ieeenat_fullname}
    \bibliography{main}
}


\end{document}